\documentclass[lettersize,journal]{IEEEtran}
\usepackage{amsmath,amsfonts}
\usepackage{algorithmic}
\usepackage{array}
\usepackage[caption=false,font=normalsize,labelfont=sf,textfont=sf]{subfig}
\usepackage{textcomp}
\usepackage{stfloats}
\usepackage{url}
\usepackage{verbatim}
\usepackage{graphicx}
\usepackage{tabularx}
\usepackage{pifont}
\usepackage{rotating}
\usepackage[table]{xcolor}

\def\BibTeX{{\rm B\kern-.05em{\sc i\kern-.025em b}\kern-.08em
    T\kern-.1667em\lower.7ex\hbox{E}\kern-.125emX}}
\usepackage{balance}

\usepackage{tikz}
\usepackage{multirow} 
\usepackage{booktabs}
\usepackage{adjustbox}
\usepackage{xcolor}
\usepackage[edges]{forest}

\usepackage{times}
\usepackage{mathptmx}
\usepackage[T1]{fontenc}
\usepackage{newtxtext}
\usepackage{newtxmath}
\renewcommand\cite{\citep}	

\definecolor{lightgreen}{rgb}{0.56, 0.93, 0.56}
\definecolor{brightlavender}{rgb}{0.75, 0.58, 0.89}
\definecolor{capri}{rgb}{0.0, 0.75, 1.0}
\definecolor{darkpastelgreen}{rgb}{0.01, 0.75, 0.24}

\definecolor{tree-level-1}{RGB}{245,20,85}
\definecolor{tree-level-2}{RGB}{246,86,118}
\definecolor{tree-level-3}{RGB}{248,177,193}
\definecolor{tree-leaf}{RGB}{176,230,198}

\definecolor{hidden-draw}{RGB}{20,68,106}
\definecolor{carminepink}{RGB}{169, 184, 198}
\definecolor{lighttealblue}{RGB}{248, 172, 140}    
\definecolor{lightplum}{RGB}{255, 136, 132}        
\definecolor{harvestgold}{RGB}{196, 151, 178}

\definecolor{my_green}{RGB}{51,102,0}
\definecolor{my_red}{RGB}{204, 0, 0}

\newcommand{\eg}{\textit{e}.\textit{g}.}
\newcommand{\etal}{\textit{et al}.}

\definecolor{hollywoodcerise}{rgb}{0.96, 0.0, 0.63}
\definecolor{lasallegreen}{rgb}{0.03, 0.47, 0.19}
\definecolor{hanpurple}{rgb}{0.32, 0.09, 0.98}
\definecolor{green(pigment)}{rgb}{0.0, 0.65, 0.31}
\usepackage[pagebackref=true,breaklinks=true,letterpaper=true,colorlinks,bookmarks=false]{hyperref}
\hypersetup{colorlinks,linkcolor={red},citecolor={hanpurple},urlcolor={red}}  

\usepackage[square, comma, sort&compress, numbers]{natbib}
\usepackage{tikz}
\usepackage{caption}
\usetikzlibrary{shapes.geometric, arrows}

\begin{document}
\title{Retrieval Augmented Generation and Understanding in Vision: A Survey and New Outlook}

\author{
Xu Zheng$^{* \dag 1,2}$,
Ziqiao Weng$^{* 1,4}$,
Yuanhuiyi Lyu$^{1}$,
Lutao Jiang$^{1}$,
Haiwei Xue$^{1,5}$, 
Bin Ren$^{2,7,8}$,
Danda Paudel$^{2}$,
Nicu Sebe$^{8}$,
Luc Van Gool$^{2,6}$,
Xuming Hu$^{\ddag 1,3}$,
\\ \vspace{5pt}
$^{1}$ HKUST(GZ),
$^{2}$ INSAIT, Sofia University “St. Kliment Ohridski”,
$^{3}$ HKUST
$^{4}$ Sichuan University,
$^{5}$ Tinghua University,
$^{6}$ ETH Zurich,
$^{7}$ University of Pisa,
$^{8}$ University of Trento,
\\  \vspace{5pt}
\small{$*$: Equal Contributions, $\dag$: Project Lead $\ddag$: Corresponding Author.} 
}


\twocolumn[{
\renewcommand\twocolumn[1][t!]{#1}%
\maketitle
\begin{center}
    \centering
    \vspace{-20pt}
    \includegraphics[width=0.9\textwidth]{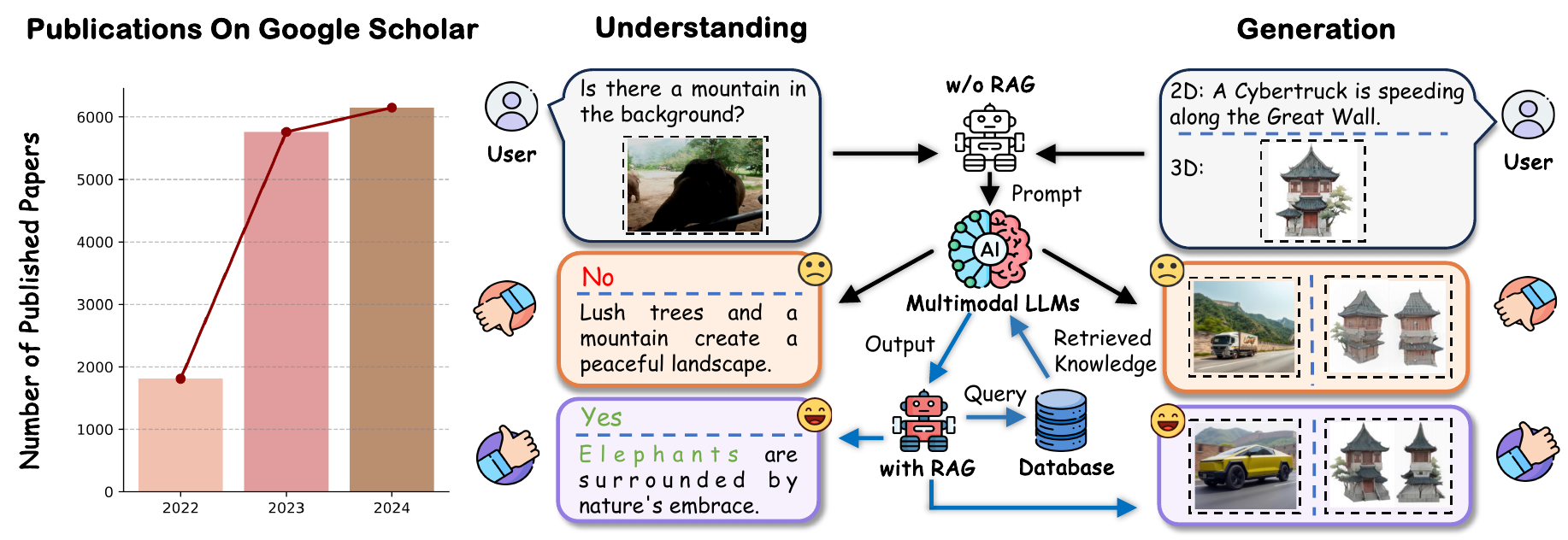}
    \captionsetup{font=small}
    \captionof{figure}{Cases from retrieval augmented visual understanding and generation.
    }
    \label{fig:cover}
\end{center}
}]

\begin{abstract}
Retrieval-augmented generation (RAG) has emerged as a pivotal technique in artificial intelligence (AI), particularly in enhancing the capabilities of large language models (LLMs) by enabling access to external, reliable, and up-to-date knowledge sources. In the context of AI-Generated Content (AIGC), RAG has proven invaluable by augmenting model outputs with supplementary, relevant information, thus improving their quality. Recently, the potential of RAG has extended beyond natural language processing, with emerging methods integrating retrieval-augmented strategies into the computer vision (CV) domain. These approaches aim to address the limitations of relying solely on internal model knowledge by incorporating authoritative external knowledge bases, thereby improving both the understanding and generation capabilities of vision models.
This survey provides a comprehensive review of the current state of retrieval-augmented techniques in CV, focusing on two main areas: (I) visual understanding and (II) visual generation. In the realm of visual understanding, we systematically review tasks ranging from basic image recognition to complex applications such as medical report generation and multimodal question answering. For visual content generation, we examine the application of RAG in tasks related to image, video, and 3D generation.
Furthermore, we explore recent advancements in RAG for embodied AI, with a particular focus on applications in planning, task execution, multimodal perception, interaction, and specialized domains. Given that the integration of retrieval-augmented techniques in CV is still in its early stages, we also highlight the key limitations of current approaches and propose future research directions to drive the development of this promising area.
Updated information about this survey can be found at \textcolor{blue}{\url{https://github.com/zhengxuJosh/Awesome-RAG-Vision}}.
\end{abstract}

\begin{table*}[h!]
\centering
\caption{Comparison of related survey papers on RAG topic.}
\setlength{\tabcolsep}{12pt}
\resizebox{\textwidth}{!}{%
\begin{tabular}{c|l|l|l|c}
\midrule
Year & Paper & Focused Areas & Main Context & GitHub \\ \midrule
2023 & Gao \etal ~\cite{gao2023retrieval} & LLMs / NLP & RAG paradigms and components & - \\ \midrule
2024 & Fan \etal ~\cite{fan2024surveyragmeetingllms} & LLMs / NLP & RA-LLMs' architectures, training, and applications & \href{https://advanced-recommender-systems.github.io/RAG-Meets-LLMs/}{link} \\ \midrule
2024 & Hu \etal ~\cite{hu2024rag} & LLMs / NLP & RA-LMs' components, evaluation, and limitations & \href{https://github.com/2471023025/RALM_Survey}{link} \\ \midrule
2024 & Zhao \etal ~\cite{zhao2024retrieval} & LLMs / NLP & challenges in data-augmented LLMs & - \\ \midrule
2024 & Gupta \etal ~\cite{gupta2024comprehensive} & LLMs / NLP & Advancements and downstream tasks of RAG & - \\ \midrule
2024 & Zhao \etal ~\cite{zhao2024retrievalaigc} & RAG in AIGC & RAG applications across modalities & \href{https://github.com/PKU-DAIR/RAG-Survey}{link} \\ \midrule
2024 & Yu \etal ~\cite{yu2024evaluation} & LLMs / NLP & Unified evalutaiton process of RAG & \href{https://github.com/YHPeter/Awesome-RAG-Evaluation}{link} \\ \midrule
2024 & Procko \etal ~\cite{procko2024graph} & Graph Learning & Knowledge graphs with LLM RAG & - \\ \midrule
2024 & Zhou \etal ~\cite{zhou2024trustworthiness} & Trustworthiness AI & Six dimensions and benchmarks about Trustworthy RAG & \href{https://github.com/smallporridge/TrustworthyRAG}{link} \\ \midrule
2025 & Singh \etal ~\cite{singh2025agentic} & AI Agent & Participles and evaluation & \href{https://github.com/asinghcsu/AgenticRAG-Survey}{link}\\ \midrule
2025 & Ni \etal ~\cite{ni2025towards} & Trustworthiness AI & Road-map and discussion & \href{https://github.com/Arstanley/Awesome-Trustworthy-Retrieval-Augmented-Generation}{link} \\ \midrule
\rowcolor{gray!10} 2025 & \textbf{\textit{Ours}} & \textbf{\textit{Computer Vision}} & \textbf{\textit{RAG for visual understanding and generation}} & \href{https://github.com/zhengxuJosh/Awesome-RAG-Vision}{link} \\
\bottomrule 
\end{tabular}
}
\label{tab:ragsurvey}
\end{table*}

\section{Introduction}
\label{sec:introduction}

\subsection{Background}
Retrieval-augmented generation (RAG) is a transformative technique in generative AI, particularly in natural language processing (NLP) and recommendation systems. It improves content quality by integrating external, up-to-date information from knowledge sources~\cite{gao2023retrieval}. Despite the impressive performance of large language models (LLMs), challenges like hallucinations, outdated knowledge, and lack of domain-specific expertise remain~\cite{gao2023retrieval}. RAG addresses these issues by supplying LLMs with relevant, retrieved factual information, enhancing model outputs.

RAG works by using a retriever to extract relevant knowledge from external databases~\cite{lewis2020retrieval}, which is then combined with the model's input to provide enriched context~\cite{yu2024visrag}. This approach is efficient, requiring minimal adaptation and often no additional training~\cite{fan2024surveyragmeetingllms}. Recent studies highlight RAG's potential, not only for knowledge-intensive tasks but also for a broad range of language-based applications~\cite{fan2024surveyragmeetingllms}, enabling more accurate and up-to-date outputs.

While traditional RAG pipelines are text-based, real-world knowledge is often multimodal, in the form of images, videos, and 3D models. This creates challenges when applying RAG to computer vision (CV). In CV, visual understanding tasks like object identification~\cite{he2016deep,lyu2024unibind}, anomaly detection~\cite{zhang2022dino}, and segmentation~\cite{ravi2024sam} require integrating external knowledge to improve accuracy~\cite{yu2024visrag}. Similarly, visual generation tasks, such as transforming textual descriptions into realistic images, can benefit from external knowledge like scene layouts, object relationships, and temporal dynamics in videos~\cite{lyu2024unibind}.

Given the complexity of visual data, RAG can significantly improve model performance. For example, scene generation models can benefit from knowledge about object interactions and spatial relationships, while image classification models can enhance accuracy by retrieving up-to-date visual references. By integrating external knowledge, RAG enhances both visual understanding and generation, helping overcome inherent challenges in vision tasks.
As shown in Figure~\ref{fig:cover}, recent research has started exploring RAG's integration into CV, aiming to improve both understanding and generation. While large vision-language models (LVLMs) have shown promise, they still face challenges with image generalization and understanding~\cite{liu2024rar}. In 3D modeling, tools like Phidias~\cite{wang2024phidias} use retrieved 3D models to guide the generation of new ones, improving quality and generalization. However, the full potential of RAG—especially in enhancing model trustworthiness, robustness, and adaptability in dynamic environments—remains underexplored, offering a valuable opportunity for future research.

\subsection{Contributions}  
This survey presents a comprehensive and systematic review of RAG techniques within the domain of computer vision, covering visual understanding, visual generation, and embodied vision. Our work synthesizes the advancements, challenges, and future directions of RAG in CV, making the following key contributions:
\textbf{(I)} We analyze the role of retrieval-augmented methods in visual understanding(Section~\ref{sec:Retrieval-Augmented Understanding in Vision}), visual generation(Section~\ref{sec:Retrieval-augmented Generation in Vision}), and embodied vision(Section~\ref{sec:embodied_ai}), focusing on image, video, and multimodal understanding tasks, as well as 3D generation, to demonstrate how RAG enhances these tasks. \textbf{(II)} We introduce a taxonomy of RAG-based techniques across various vision tasks, highlighting key contributions and differences, and comparing methods in pattern recognition, medical vision, and video analysis. (Section~\ref{sec:Retrieval-Augmented Understanding in Vision}) \textbf{(III)} We identify key limitations in current RAG applications, such as retrieval efficiency, modality alignment, computational cost, and domain adaptation, and discuss the challenges hindering wider adoption. (Section~\ref{sec:insights_outlook}) \textbf{(IV)} We propose future research directions to advance RAG in computer vision, focusing on real-time retrieval optimization, cross-modal retrieval fusion, privacy-aware retrieval, and retrieval-based generative modeling, highlighting new opportunities for exploration.(Section~\ref{sec:insights_outlook}) \textbf{(V)} We extend RAG applications beyond text-based retrieval, exploring multimodal frameworks for enhancing vision models, and discuss its potential for embodied AI, 3D content generation, and multimodal learning in robotics, autonomous driving, and real-world decision-making.(Section~\ref{sec:insights_outlook})

This survey serves as a foundational resource, synthesizing existing works and offering insights into the future development of retrieval-augmented techniques in computer vision.

\subsection{Related Works} Table~\ref{tab:ragsurvey} summarizes key contributions in the field of RAG, primarily focusing on language models (LLMs) and the integration of external knowledge across modalities. Gao \etal~\cite{gao2023retrieval} provide an overview of RAG paradigms, highlighting how external knowledge enhances LLMs. Fan \etal~\cite{fan2024surveyragmeetingllms} expand on this by discussing RA-LLMs, their architectures, training strategies, and applications. Hu \etal~\cite{hu2024rag} offer a comprehensive evaluation of Retrieval-Augmented Language Models (RALMs), addressing their components, limitations, and areas for improvement. Zhao \etal~\cite{zhao2024retrieval} identify challenges in data-augmented LLMs, focusing on retrieval quality and data integration. Gupta \etal~\cite{gupta2024comprehensive} review advancements in RAG and explore downstream tasks, shedding light on its evolving role in language processing. Zhao \etal~\cite{zhao2024retrievalaigc} extend RAG's scope to AI-Generated Content (AIGC), emphasizing its cross-modal applications. Yu \etal~\cite{yu2024evaluation} propose a unified evaluation framework for RAG, standardizing performance metrics across tasks. Procko \etal~\cite{procko2024graph} investigate the integration of knowledge graphs with LLM-based RAG systems. Zhou \etal~\cite{zhou2024trustworthiness} focus on trustworthiness in RAG, advocating for robust benchmarks to ensure reliable performance. Singh \etal~\cite{singh2025agentic} and Ni \etal~\cite{ni2025towards} explore agentic RAG and trustworthiness, respectively, offering frameworks for evaluating and discussing future directions.

\begin{figure*}[th!]
    \centering
    \includegraphics[width=0.95\textwidth]{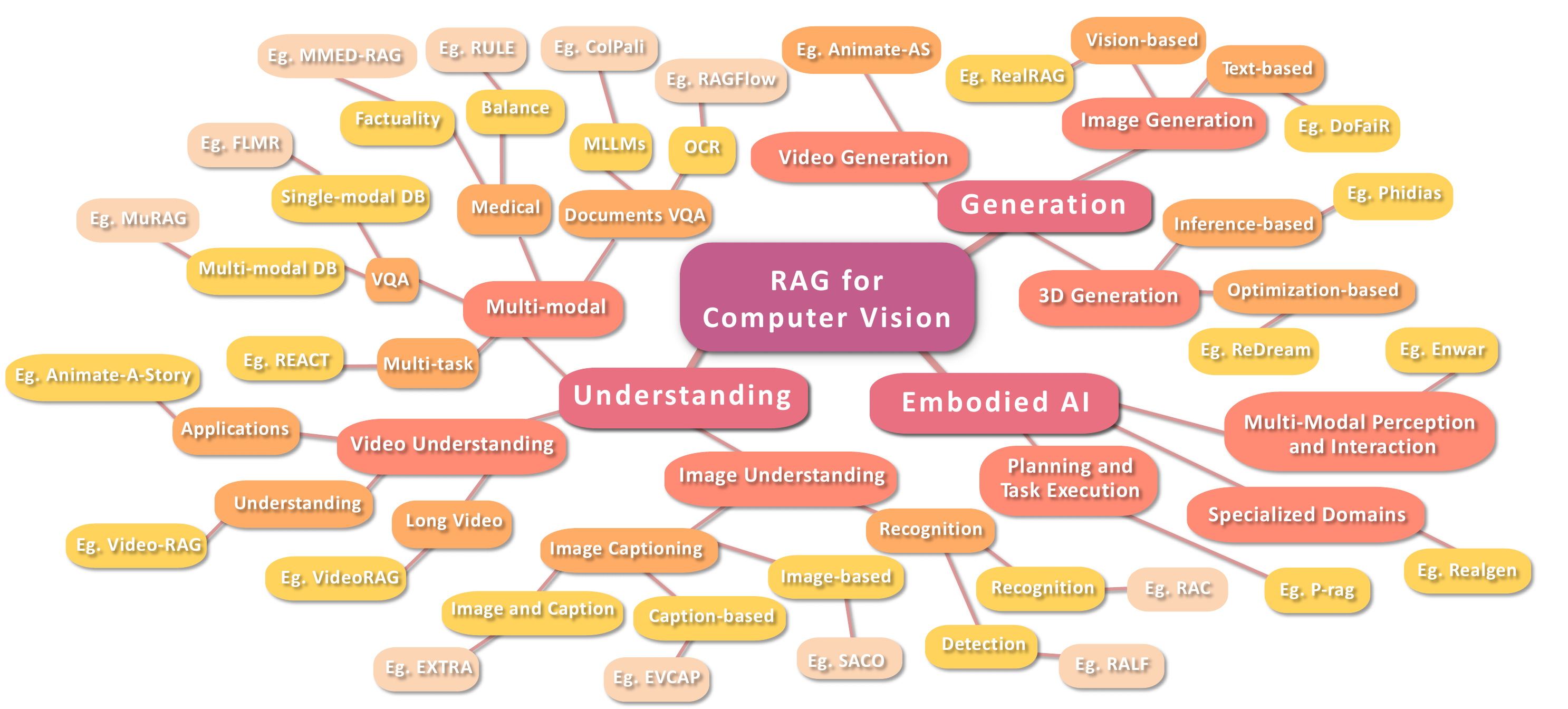}
    \caption{
    The main categorization of this survey. 
    }
    \label{fig:article_mindmap}
\end{figure*}

Building on these foundational works, our research is the \textbf{\textit{first}} to focus on RAG in computer vision (CV). 
We provide a concise review of RAG for visual understanding and generation, emphasizing its potential and challenges. 
Our work extends RAG from language models to visual tasks, offering insights into its impact on CV and paves the way for future research by exploring methods that combine retrieval-based techniques with visual perception, aiming to improve performance and real-world applicability.
The systemic overview of our paper is in Figure~\ref{fig:article_mindmap}.

\section{Retrieval-Augmented Understanding in Vision}
\label{sec:Retrieval-Augmented Understanding in Vision}
\subsection{Image Understanding}
\label{subsec:understanding_image}

\begin{figure*}[th!]
    \centering
    \includegraphics[width=0.95\textwidth]{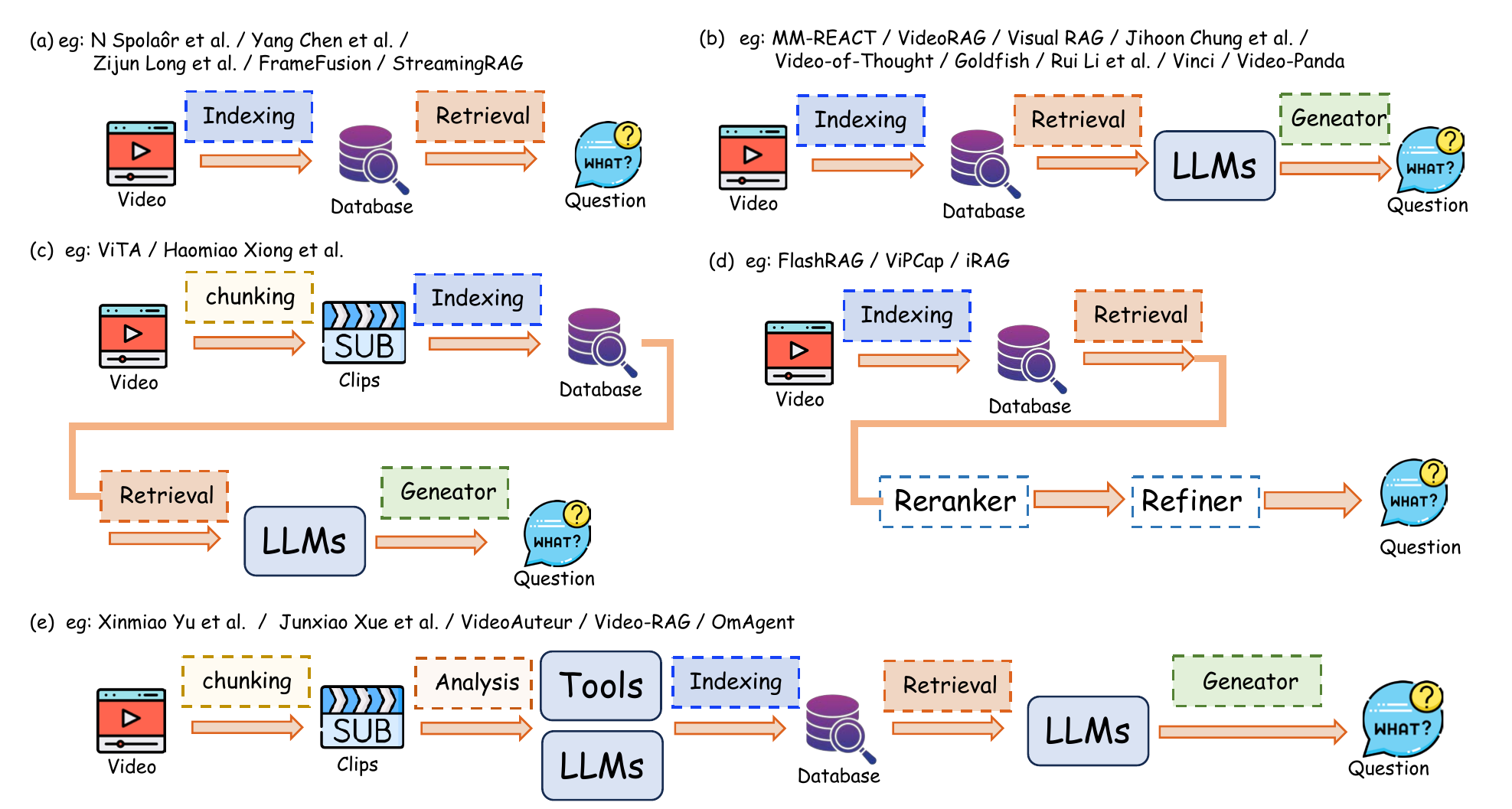}
    \caption{
    Five different RAG for video retrieval and understanding pipelines. 
    }
    \label{fig:video_rag_pipeline}
\end{figure*}

\begin{table*}[h!]
\centering
\caption{Video Retrieval and Understanding Methods}
\label{tab:video_retrieval}
\resizebox{\textwidth}{!}{
\begin{tabular}{c|c|c|c|c}
\toprule
\textbf{Method} & \textbf{Retrieval Metric} & \textbf{Augmentation Method} & \textbf{Video Specific Operation} & \textbf{Video Corpus} \\ \midrule
VideoRAG~\cite{jeong2025videorag} & Cosine similarity  & Concatenation & Adaptive frame selection & Howto100m~\cite{miech2019howto100m} \\ \midrule
ViTA~\cite{Arefeen2024ViTA} & MLLMs captions & Textual description & Video clips & StreetAware~\cite{piadyk2023streetaware} \& Tokyo MODI~\cite{kossmann2023extract} \\ \midrule
OmAgent~\cite{zhang2024omagent} & MLLMs captions & Scene descriptions & Detection / Face recognition & Sok-bench~\cite{wang2024sok} \\ \midrule
StreamingRAG~\cite{sankaradas2025streamingragrealtimecontextualretrieval} & MLLMs captions & Knowledge base & Temporal context identifier & - \\
\bottomrule
\end{tabular}
}
\end{table*}

\subsubsection{Pattern Recognition}

Real-world data follows a long-tailed distribution, making it impractical to store every visual cue due to hardware constraints. To address this, retrieval-augmented methods enhance pattern recognition tasks like segmentation and detection. \textit{RAC}~\cite{long2022retrieval} augments standard image classification pipelines with a retrieval module, incorporating relevant external information to improve recognition accuracy.
Beyond classification, retrieval-augmented methods benefit detection tasks, particularly in open-set scenarios where traditional object detection struggles with closed category sets. Expanding the vocabulary and incorporating diverse concepts significantly enhance open-vocabulary detection. \textit{RALF}~\cite{kim2024retrieval} retrieves related classes and verbalizes concepts, enriching visual features with semantically meaningful context.
For segmentation, \cite{zhao2024retrieval} utilizes DINOv2 features as queries to retrieve similar samples from a limited annotated dataset, encoding them into a memory bank. Using SAM 2’s memory attention mechanism, the model conditions segmentation predictions on these stored samples, improving accuracy.

\subsubsection{Image Captioning} 

Image captioning involves generating textual descriptions for images. Retrieval-augmented approaches fall into three categories: (1) retrieving similar or style-aware images before caption generation, (2) retrieving related captions, and (3) retrieving both image and caption embeddings.
In the first category, SACO~\cite{zhou2023style} addresses the relationship between linguistic style and visual content using object-, RoI-, and triplet-based retrieval to capture style-relevant visual features. The second category includes EVCAP~\cite{li2024evcap}, which retrieves object names from an external visual memory to prompt LLMs for caption generation.
In the third category, EXTRA~\cite{ramos2023retrieval} encodes both input images and retrieved captions to enhance textual context. SAMLLCAP~\cite{li2024understanding} explores model robustness by training with sampled retrieved captions from a larger pool, demonstrating that dynamic retrieval improves captioning resilience over fixed top-k selection.

\subsection{Video Understanding}
\label{subsec:understanding_video}
\subsubsection{Video Understanding}  

As shown in Table~\ref{tab:video_retrieval}, video retrieval and understanding have advanced significantly, driven by multimodal architectures and retrieval-augmented generation (RAG). This section examines key technical developments and their impact.  
MM-REACT~\cite{yang2023mm} pioneered a multi-expert system integrating ChatGPT with vision specialists via prompting. This foundation led to modular frameworks like FlashRAG~\cite{jin2024flashrag}, enhancing RAG efficiency, and domain-specific adaptations such as VideoRAG~\cite{jeong2025videorag} and Visual RAG~\cite{bonomo2025visual} for video understanding and visual knowledge expansion. However, balancing computational efficiency with comprehensive understanding remains a challenge.  
Practical implementations have refined video understanding. ViTA~\cite{Arefeen2024ViTA} optimized a video-to-text pipeline for production, while OmAgent~\cite{zhang2024omagent} employed a divide-and-conquer strategy for complex tasks. 
Recent work has improved retrieval accuracy and efficiency. Enhanced Multimodal RAG-LLM~\cite{xue2024enhanced} boosted visual question answering, while ViPCap~\cite{kim2024vipcap} demonstrated the advantages of integrating video and text retrieval.

\subsubsection{Long Video Retrieval and Understanding}  
Long video understanding poses significant challenges, necessitating innovative solutions across multiple technical dimensions. A key area of advancement is real-time processing and streaming capabilities, with various approaches addressing different facets of this problem. StreamingRAG~\cite{sankaradas2025streamingragrealtimecontextualretrieval} pioneered real-time contextual retrieval, while Vinci~\cite{huang2024vinci} demonstrated its practical utility in embodied assistance. Xiong \etal~\cite{xiong2025streaming} further advanced the field by enhancing memory-driven streaming comprehension. Despite these advancements, maintaining an optimal balance between processing speed and comprehension accuracy remains a critical challenge.  
To address specific aspects of long video understanding, specialized optimization techniques have been proposed. Video-RAG~\cite{luo2024video} introduced visually-aligned retrieval augmentation to enhance representation learning, while iRAG~\cite{arefeen2024irag} developed incremental processing techniques to improve computational efficiency. These methods contribute to advancing long video analysis; however, challenges related to maintaining consistency over extended sequences and handling complex temporal dependencies persist.  

As illustrated in Figure~\ref{fig:video_rag_pipeline}, five distinct RAG pipelines for video retrieval and understanding are categorized based on their architectural approaches:

(a) Basic Retrieval Pipeline (e.g., N Spolaôr et al.~\cite{spolaor2020systematic}, Yang Chen et al.~\cite{chen2024large}) employs direct indexing and retrieval from video databases without additional processing modules.

(b) LLM-Augmented Pipeline (e.g., MM-REACT~\cite{yang2023mm}, VideoRAG~\cite{jeong2025videorag}) integrates large language models (LLMs) as generators to enhance retrieval outputs through semantic understanding.

(c) Chunking-Based Pipeline (e.g., ViTA~\cite{Arefeen2024ViTA}, Haomiao Xiong et al.~\cite{xiong2025streaming}) introduces video chunking and sub-clip processing before retrieval, enabling fine-grained analysis of segmented content.

(d) Multi-Stage Refinement Pipeline (e.g., FlashRAG~\cite{jin2024flashrag}, ViPCap~\cite{kim2024vipcap}) incorporates rerankers and refiners to iteratively optimize retrieval results through post-processing stages.

(e) Tool-Enhanced Pipeline (e.g., Xinmiao Yu et al.~\cite{yu2024cross}, Video-RAG~\cite{luo2024video}) combines chunking with specialized analysis tools and multi-step LLM processing, enabling sophisticated multimodal reasoning.

These pipelines demonstrate evolving methodologies in video RAG systems, ranging from basic retrieval to complex multimodal architectures leveraging LLMs and domain-specific tools.


\subsubsection{Other Retrieval and Application Tasks}  

The application of retrieval-augmented approaches has expanded into diverse domains, each presenting distinct challenges and opportunities. This section examines key advancements across various application areas, highlighting technical innovations and persistent challenges.  
Retrieval-augmented methods have played a crucial role in enhancing storytelling and visualization. Animate-A-Story~\cite{he2023animate} demonstrated the potential of retrieval-augmented video generation for producing coherent narratives, while Dialogue Director~\cite{zhang2024dialogue} facilitated the transformation of dialogue-centric scripts into visual storyboards. Despite these advancements, maintaining narrative consistency and visual coherence across generated content remains a significant challenge.  
Ensuring the reliability and security of video understanding systems has become increasingly important. Wen \etal~\cite{wen2024ensemble} introduced ensemble-based approaches for short-form video quality assessment using multimodal LLMs. Concurrently, Fang \etal~\cite{fang2025retrievals} uncovered security vulnerabilities in retrieval-augmented diffusion models through contrastive backdoor attacks, emphasizing the necessity of robust security measures in retrieval-augmented systems.  
The versatility of retrieval-augmented approaches has been demonstrated in various specialized domains. Hong \etal~\cite{hong2024free} advanced free-viewpoint human animation through pose-correlated reference selection, while Luo \etal~\cite{luo2025graphbasedcrossdomainknowledgedistillation} tackled cross-domain person retrieval using graph-based knowledge distillation. Video moment retrieval has seen significant improvements, with Xu \etal proposing zero-shot retrieval~\cite{xu2025zero} and multimodal fusion techniques~\cite{xu2024multi}. Additionally, Chen \etal~\cite{chen2024human} contributed to dataset expansion through human-guided image generation.  


\begin{table*}[t!]
\caption{Comparison of multimodal Understanding Benchmarks. This table presents a comprehensive overview of various benchmarks categorized by their primary evaluation focus. The scale information is approximate and based on reported numbers in the papers. Some benchmarks may use multiple metrics or combined datasets.}
\label{tab:multimodal-understanding-benchmarks}
\centering
\setlength{\tabcolsep}{2pt}
\resizebox{\textwidth}{!}{
\renewcommand{\arraystretch}{1.1}
\begin{tabular}{llllll}
\toprule
Category & Benchmark & Year & Scale & Key Features & Evaluation Focus \\
\midrule
\multirow{7}{*}{\textbf{\begin{tabular}{@{}c@{}}Traditional\\ VQA\end{tabular}}} & VQA v2~\cite{goyal2017makingvvqa} & 2017 & 250K images, 1.1M QA pairs & Balanced visual-linguistic understanding & Open-ended answer generation \\
& Vizwiz-VQA~\cite{gurari2018vizwiz} & 2018 & 20,523 question pairs & Queries for visually impaired users & Robustness and real-world adaptability \\
& OK-VQA~\cite{marino2019okvqa} & 2019 & 14k QA pairs & External knowledge-based VQA & Knowledge retrieval and reasoning \\
& A-OKVQA~\cite{10.1007/978-3-031-20074-8_9} & 2022 & 25k QA pairs & Extended OK-VQA & Complex reasoning \\
& ScienceQA~\cite{lu2022learnexplainmultimodal} & 2022 & 21k multiple-choice queries & 3 Major domains, 379 skills & Interdisciplinary Integrated Reasoning \\
& OVEN~\cite{hu2023opendomain} & 2023 & select among 6M Wikipedia entities & knowledge-intensive QA & Deep integration of visual semantics \\
& InfoSeek~\cite{chen2023pretrainedvision} & 2023 & 8.9K human-written IQA & Information-seeking VQA & Fine-grained visual attributes \\
\midrule
\multirow{3}{*}{\textbf{\begin{tabular}{@{}c@{}}Traditional\\ Doc-VQA\end{tabular}}} & M3DocVQA~\cite{cho2024m3docrag} & 2024 & 2,441 multi-hop queries & Multi-page, multi-document retrieval & Open-domain multi-hop QA \\
& ViDoRe~\cite{faysse2024colpali} & 2024 & Wide coverage of domains & Document visual context retrieval & Visual element sensitivity \\
& MMDocIR~\cite{dong2025mmdocir} & 2025 & 1,685 QA pairs, 313 documents & Long-document support & Long-context understanding \\
\midrule
\multirow{3}{*}{\textbf{\begin{tabular}{@{}c@{}}Traditional\\ Fundamental Vision Tasks\end{tabular}}} & MS-COCO~\cite{10.1007/978-3-319-10602-1_48} & 2014 & 330K images (\textgreater{}200K labeled) & Cross-task & Image captioning quality \\
& LVIS~\cite{gupta2019lvisdataset} & 2019 & 164k images, 1,203 categories & Zero-shot recognition & Rare category recognition precision \\
& V3Det~\cite{wang2023v3det} & 2023 & 13,204 categories & Ultra-fine-grained object detection & Large-scale category generalization \\
\midrule
\multirow{4}{*}{\textbf{General}} & ELEVATER~\cite{li2022elevater} & 2022 & 20 image classification datasets & Unified evaluation toolkit & Cross-task generalization \\
& MMBench~\cite{liu2024mmbench} & 2023 & 2,974 multiple-choice queries & 20 Ability dimensions & Multi-dimensional capability \\
& SEED-Bench~\cite{LiGGWWZS24} & 2023 & 19k multiple-choice queries & 12 Evaluation dimensions & Dynamic scene understanding \\
& MMStar~\cite{chen2024rightway} & 2024 & 1,500 samples & High-quality review & Comprehensive cognitive capability \\
\midrule
\multirow{3}{*}{\textbf{\begin{tabular}{@{}c@{}}Technical\\ RAG\end{tabular}}} & Visual-RAG~\cite{wu2025visualrag} & 2024 & 400 Qs, 103824 images & Text-image joint retrieval & Quality and multimodal alignment \\
& MRAG-Bench~\cite{hu2025mragbench} & 2024 & 16,130 images and 1,353 queries & Visual-centered retrieval enhancement & Visual semantic retrieval effectiveness \\
& REAL-MM-RAG~\cite{wasserman2025realmmra} & 2025 & 8000 pages and 5000 queries & Real-world multimodal retrieval & Multimodal retrieval \\
\midrule
\multirow{2}{*}{\textbf{\begin{tabular}{@{}c@{}}Technical\\ Hallucination\end{tabular}}} & Pope~\cite{li2023evaluatingobject} & 2023 & Adaption from MSCOCO & Detection object hallucination & Hallucination suppression \\
& RAG-Check~\cite{mortaheb2025rag} & 2025 & 121,000 samples & Multi-source integration & Relevancy score \\
\bottomrule
\end{tabular}}
\end{table*}
\subsection{Multimodal Understanding}
\label{subsec:understanding_multi-modal}

\begin{figure*}[t!]
    \centering
    \includegraphics[width=1\textwidth]{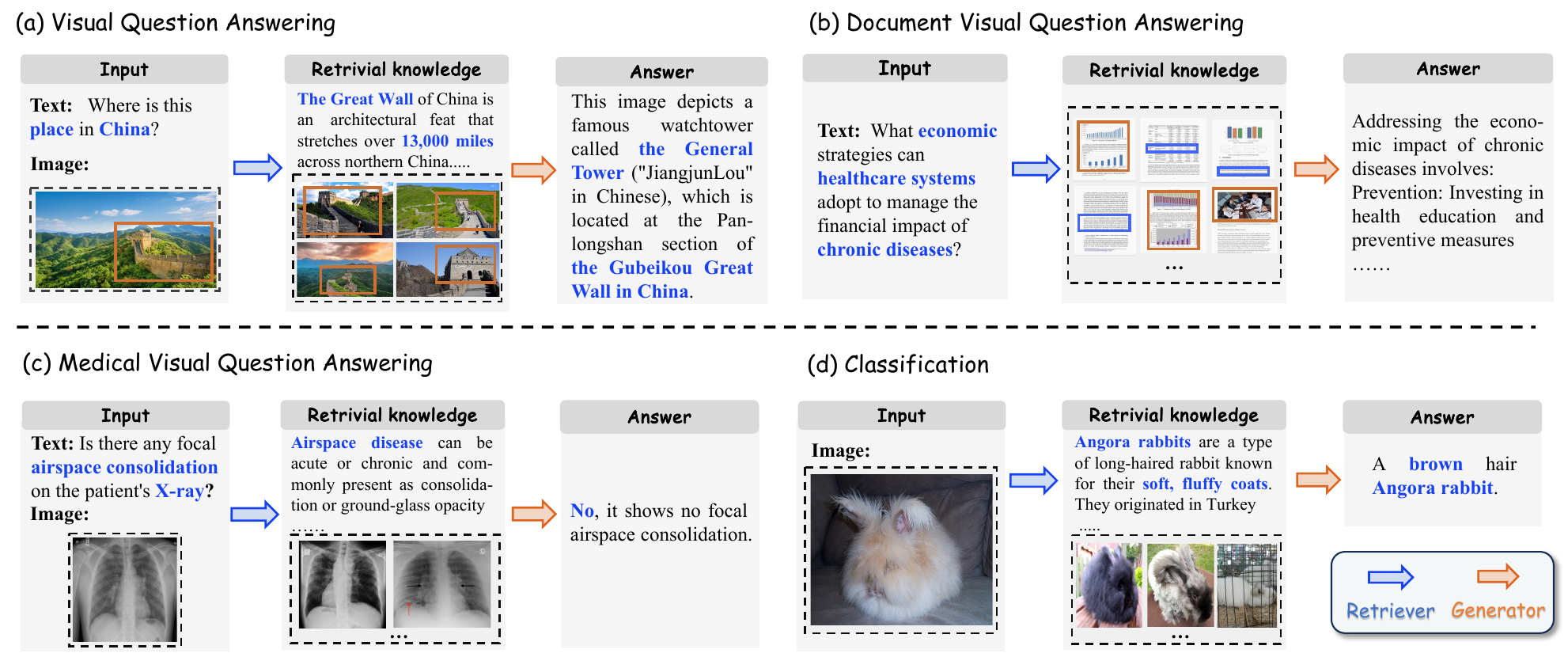}
    \caption{
    Frameworks for different multimodal RAG application.
    }
    \label{fig:MMRAU_application}
\end{figure*}

\begin{figure}[t!]
    \centering
    \includegraphics[width=\linewidth]{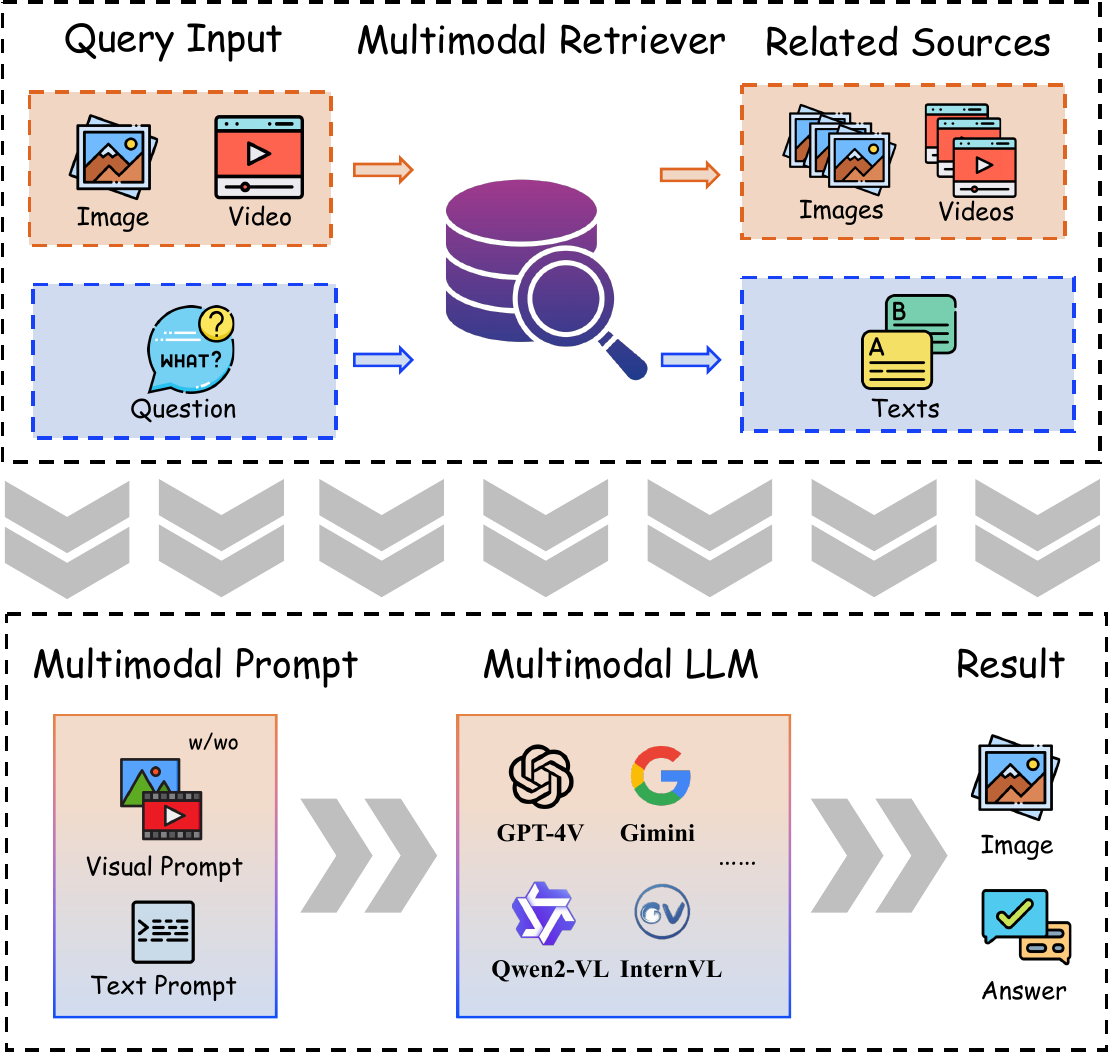}
    \caption{
    The general framework of multimodal RAG tasks.
    }
    \label{fig:multi_modal_task_framework}
\end{figure}

Though the application of RAG in the NLP domain has been extensively explored in various downstream tasks~\cite{fan2024surveyragmeetingllms}, its adaptation within the multimodal learning domain remains in its early stages. Pioneering works such as MuRAG~\cite{chen2022murag} and REACT~\cite{liu2023learning} have highlighted the potential of multimodal retrieval and reasoning mechanisms in enhancing the inferential capabilities of vision-language models. 
As shown in Figure~\ref{fig:MMRAU_application}, the applications of RAG in multimodal settings can be primarily categorized into three major directions according to the downstream tasks: visual question answering, document understanding, medical visual question answering, and general-purpose multi-task integration with classification as an example. The overall architecture of a multimodal retriever is illustrated in Figure~\ref{fig:multi_modal_task_framework}, which provides a comprehensive view of the retrieval pipeline.

\begin{figure*}[ht!]
    \centering
    \includegraphics[width=0.85\textwidth]{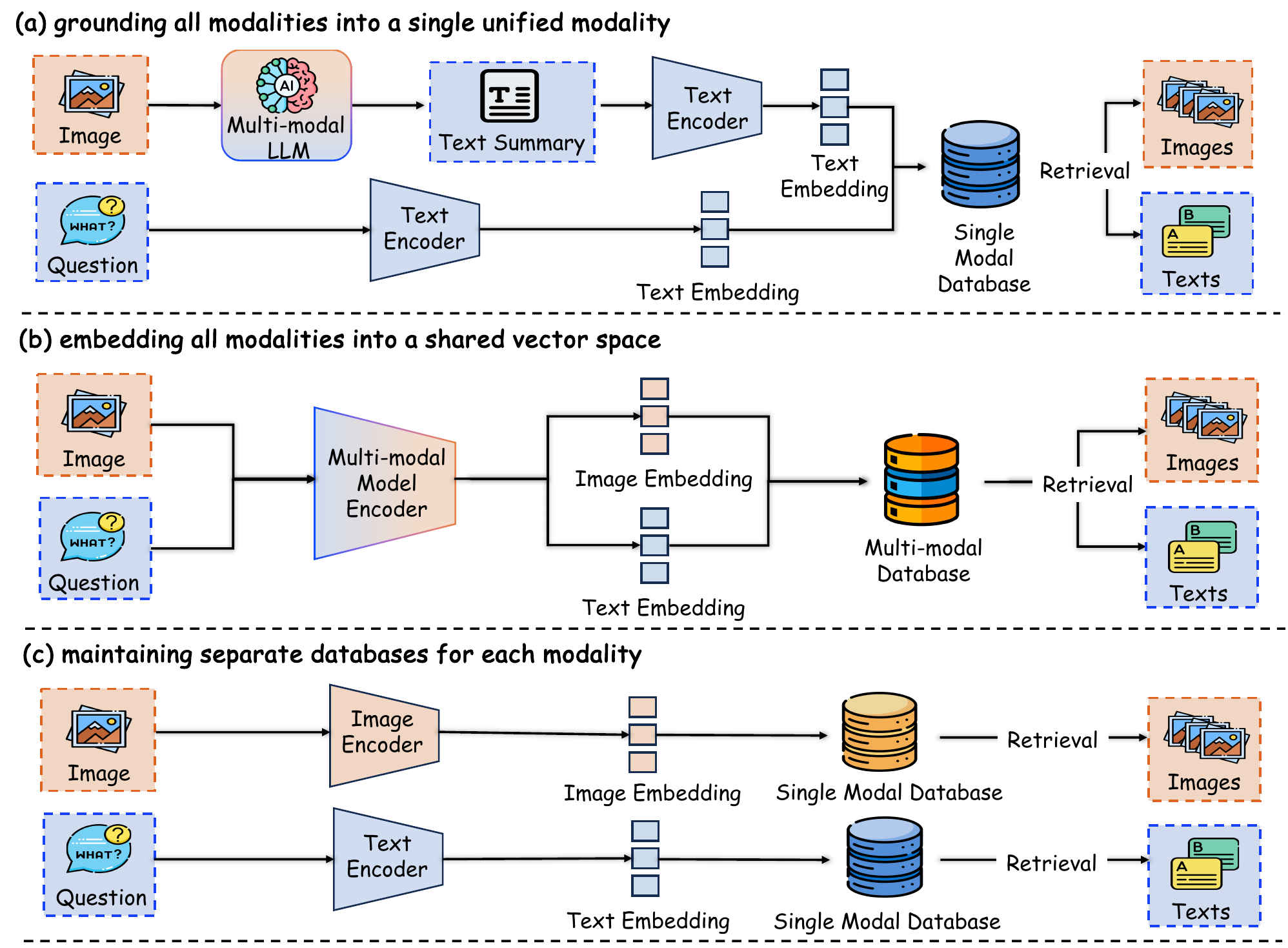}
    \caption{
    Three different multimodal RAG pipelines. \textbf{(a)} grounding all modalities into a single unified modality; \textbf{(b)} embedding all modalities into a shared vector space; \textbf{(c)} maintaining separate databases for each modality.
    }
    \label{fig:multimodal_retriever_pipeline}
\end{figure*}
\subsubsection{Benchmark Datasets}

In multimodal retrieval-augmented understanding (MMRAU), benchmark datasets are essential for evaluating model performance, advancing technology, and standardizing research. As shown in Table~\ref{tab:multimodal-understanding-benchmarks}, we categorize these datasets into three types: traditional evaluation datasets, general-purpose benchmarks, and technology-focused datasets. This classification captures the diverse requirements of various applications and research directions, highlighting differences in scale, characteristics, and evaluation priorities.
Traditional datasets can be categorized into Visual Question Answering (VQA), Document VQA (DocVQA), and fundamental vision tasks. VQA datasets emphasize image-text interaction. VQA v2~\cite{goyal2017makingvvqa} evaluates open-ended answer generation, while Vizwiz-VQA~\cite{gurari2018vizwiz} targets visually impaired users. OK-VQA~\cite{marino2019okvqa} and A-OKVQA~\cite{10.1007/978-3-031-20074-8_9} introduce external knowledge, and datasets like ScienceQA~\cite{lu2022learnexplainmultimodal} and OVEN~\cite{hu2023opendomain} demand cross-disciplinary reasoning and deep visual-semantic integration. These serve as comprehensive multimodal QA benchmarks.
DocVQA datasets focus on visual-textual integration in documents. M3DocVQA~\cite{cho2024m3docrag} supports multi-page, multi-document retrieval, ViDoRe~\cite{faysse2024colpali} emphasizes document context retrieval, and MMDocIR~\cite{dong2025mmdocir} addresses long-document comprehension—filling gaps in traditional VQA.
Fundamental vision task datasets support core computer vision benchmarks. MS-COCO~\cite{lin2014microsoft} is used for image captioning, LVIS~\cite{gupta2019lvisdataset} for zero-shot recognition, and V3Det~\cite{wang2023v3det} for ultra-fine-grained object detection. While all focus on images, they vary in task complexity, class balance, and generalization.

The second category consists of general-purpose datasets designed to assess a model’s comprehensive multimodal capabilities across various tasks. ELEVATER~\cite{li2022elevater} provides a unified evaluation framework, focusing on image classification and object detection, while MMBench~\cite{liu2024mmbench} introduces a fine-grained capability-based assessment. SEED-Bench~\cite{LiGGWWZS24} and MMStar~\cite{chen2024rightway} set higher benchmarks in dynamic scene understanding and integrated cognitive abilities. However, current general-purpose datasets face challenges such as insufficient evaluation dimensions and imbalanced task coverage. Future improvements are necessary to better align with the diverse requirements.
The final category encompasses datasets with a strong emphasis on specific technological aspects. We divide it into RAG-related datasets and hallucination detection datasets. RAG datasets focus on retrieval-augmented generation tasks. Visual-RAG~\cite{wu2025visualrag} centers on joint text-image retrieval, emphasizing retrieval quality and multimodal alignment. MRAG-Bench~\cite{hu2025mragbench} shifts the focus toward vision-centric retrieval enhancement. REAL-MM-RAG~\cite{wasserman2025realmmra} extends to real-world multimodal retrieval tasks. Hallucination datasets aim to evaluate a model’s ability to suppress hallucinations during the generation process. Pope~\cite{li2023evaluatingobject}, adapted from MS-COCO~\cite{lin2014microsoft}, is specifically designed for detecting object hallucinations. RAG-Check~\cite{mortaheb2025rag} further expands to multi-source information integration, focusing on relevance scoring.

\subsubsection{Visual Question Answering (VQA)}

The integration of textual and visual information has become a key focus in enhancing model performance for the visual question answering (VQA) task. However, many existing methods predominantly emphasize the knowledge embedded in the language modality, often overlooking the rich and valuable information provided by other modalities, such as images~\cite{Goyal2019VQA}.

To address this limitation, leveraging multimodal data from comprehensive multimodal databases presents a promising solution. The MuRAG framework~\cite{chen2022murag} was the first to construct such a database, combining retrieval and language augmentation techniques to generate high-quality outputs. This approach, which integrates multimodal retrieval, is categorized as pipeline (b), as illustrated in Figure~\ref{fig:multimodal_retriever_pipeline}. Similarly, UDKAG~\cite{li2024search} adopts a similar methodology but places a stronger emphasis on the recency of knowledge by incorporating web-retrieved data during inference.

In contrast, pipeline (a) in Figure~\ref{fig:multimodal_retriever_pipeline} generates image representations through image-to-text conversion, encoding visual content into descriptive text before further processing. While this method can capture certain aspects of the image, it often fails to preserve the full richness and complexity of visual information. This shortcoming arises from the inherent limitations of converting images into text, where crucial visual details can be lost or inaccurately represented.

To overcome these challenges, FLMR~\cite{lin2023finegrained} introduces an advanced approach that combines late interaction with multi-dimensional representations to more effectively capture fine-grained, cross-modal relevance between text and images. Unlike image-to-text conversion, FLMR fosters a more nuanced interaction between the modalities, preserving intricate visual features that are essential for accurate reasoning. By utilizing multi-dimensional representations, FLMR enhances the model's capacity to capture deeper, more detailed information from both image and text, thereby improving cross-modal understanding and performance in tasks requiring sophisticated integration of visual and textual data. This method offers a compelling alternative to traditional image-to-text approaches, addressing the issue of incomplete image representations and improving the quality of multimodal interactions.

Recent advancements in multimodal retrieval-augmented generation (RAG) for VQA, such as RMR~\cite{tan2024retrieval} and RagLLaVA~\cite{chen2024mllm}, have further refined this approach. These methods are based on pipeline (c) in Figure~\ref{fig:multimodal_retriever_pipeline}, with RMR~\cite{tan2024retrieval} enhancing the model’s reasoning capabilities by introducing in-context learning (ICL) within the multimodal RAG framework. Meanwhile, RagLLaVA~\cite{chen2024mllm} improves model robustness by incorporating knowledge-enhanced re-ranking and noise injection during training.

\begin{table*}[h!]
\centering
\setlength{\tabcolsep}{16pt}
\caption{Summary of Document Understanding Methods with Core Techniques and Retrieval Models}
\resizebox{\textwidth}{!}{  
\begin{tabular}{l|p{5cm}|p{5cm}}
\midrule
\textbf{Method} & \textbf{Document Processing Methods} & \textbf{Retrieval Methods / Models} \\ \midrule
DocVQA~\cite{mathew2021docvqa} & Document page segmentation & Text-based retriever with MLLMs \\ \midrule
OCR-based RAG~\cite{ragflow} & OCR-based text summarization & OCR and text-based retriever \\ \midrule
ColPali~\cite{faysse2024colpali} & Embedding images of documents & VLM-based retriever \\ \midrule
VisRAG~\cite{yu2024visrag} & Treating documents as image & VLM-based retriever \\ \midrule
M3DocRAG~\cite{cho2024m3docrag} & Multi-page, multi-document RAG & Multimodal retriever \\ \midrule
Beyond Text~\cite{riedler2024text} & Industrial RAG applications & Multimodal retriever \\ \bottomrule
\end{tabular}
}
\end{table*}

\subsubsection{Document Understanding}

Document visual question answering (DocVQA) is a specialized application of visual question answering (VQA) that focuses on answering textual queries by interpreting the information contained within documents)~\cite{mathew2021docvqa}. 
Currently, the majority of DocVQA applications are based on multimodal large language models(MLLMs)~\cite{ye2023mplugdoc} by splitting documents into individual page images, corresponding to pipeline (b)~in Figure~\ref{fig:multimodal_retriever_pipeline}. While this architecture offers simplicity, it imposes high demands on the model~\cite{cho2024m3docrag}, requiring an embedding system capable of accurate different modalities integration. This method also faces difficulty in handling long, multi-page documents, particularly in maintaining the relationships between pages.

Methods in industrial scenarios often focus on OCR-based RAG~\cite{ragflow}, which first summarizes the image content into a text summary, corresponding to pipeline (a)~in Figure~\ref{fig:multimodal_retriever_pipeline}. This approach relies on the direct retrieval and generation of text blocks, which effectively improves the quality of document understanding. However, it faces the potential loss of critical visual information in conversion, such as graphics.

To address these challenges, ColPali~\cite{faysse2024colpali} utilizes vision-language models to generate high-quality contextualized embeddings from document pages, significantly enhancing performance in visually rich document retrieval. Unlike traditional OCR methods, this approach focuses on improving multimodal retrieval and falls under pipeline (b) in Figure~\ref{fig:multimodal_retriever_pipeline}. However, it primarily addresses innovations in the retrieval component, without emphasizing the generation aspect.

Building on ColPali~\cite{faysse2024colpali}, VisRAG~\cite{yu2024visrag} and M3DocRAG~\cite{cho2024m3docrag} integrate the generation module and perform end-to-end RAG evaluations. M3DocRAG~\cite{cho2024m3docrag}, in particular, emphasizes handling multi-page, multi-document scenarios. Similarly, Beyond Text~\cite{riedler2024text} explores the application of multimodal RAG in industrial environments, highlighting the promising future prospects of this approach.

\subsubsection{General-purpose Multi-task Integration}

Most multimodal RAG methods are designed for single tasks, such as VQA~\cite{chen2024mllm} or DocVQA~\cite{chen2022murag}. Due to the high cost of pretraining and extra parameter overhead~\cite{wang2024charxiv}, multi-task RAG for vision-language models (VLMs) remains underexplored.
Current multi-task setups primarily adopt pipeline (b) in Figure~\ref{fig:multimodal_retriever_pipeline}, with few using pipeline (c). Pipeline (b) benefits from a unified vector space, enabling efficient cross-modal comparison and reducing integration complexity. In contrast, pipeline (c) handles heterogeneous data types, increasing database management complexity.

ReVeaL~\cite{hu2023reveal} first introduces a knowledge-aware self-supervised learning approach, effectively integrating world knowledge into the model. Similarly, REAVL~\cite{rao2023retrieval} explores the role of retrieving world knowledge from knowledge graphs. Those methods enhance the multimodal learning capabilities and benefiting tasks such as VQA and image captioning.

For classic downstream tasks in computer vision, such as classification, detection, and segmentation, REACT~\cite{liu2023learning} pioneered task-specific model enhancement by training new blocks without altering original weights. RAVEN~\cite{rao2024raven}, on the other hand, utilizes a multi-task learning framework, simultaneously handling multiple tasks. Frameworks such as SURf~\cite{sun2024surf} and RoRA-VLM~\cite{qi2024roravlm} complement this by selectively utilizing retrieved information to enhance the model's robustness against irrelevant or misleading data. Similarly, RAR~\cite{liu2024rar} optimizes performance through retrieval and ranking with MLLM for fine-grained knowledge tasks.

\subsubsection{RAG in Medical Domain}
Medical large vision-language models (Med-LVLMs) have shown significant promise in advancing interactive and intelligent diagnosis~\cite{xia2024mmed, xia2024rule}. These models integrate medical knowledge from visual inputs (such as images, videos, and scans) and textual data, enabling them to understand complex medical contexts and assist in decision-making processes. However, despite their potential, current Med-LVLMs still face significant challenges, particularly with hallucinations—generating non-factual or misleading medical responses—which undermines their reliability in critical healthcare applications.

To address these limitations, RAG has emerged as a viable solution. By integrating external knowledge sources during the inference process, RAG enhances the accuracy and reliability of Med-LVLMs. This approach works by retrieving relevant, domain-specific knowledge from external databases, such as medical literature or image repositories, and incorporating it into the model's decision-making.
MMED-RAG~\cite{xia2024mmed} proposes a versatile multimodal RAG system specifically designed to improve the factuality of Med-LVLMs. This system integrates a domain-aware retrieval mechanism to selectively extract relevant medical knowledge from a structured knowledge base, ensuring the generated responses are both accurate and contextually relevant. 

Further, the process of retrieval can be augmented by advanced mechanisms that refine the selection of context, RULE~\cite{xia2024rule} addresses the challenges of directly applying RAG to Med-LVLMs, such as limitations in the number of retrieved contexts and the potential over-reliance on external sources. By introducing a more adaptive retrieval and context selection process, these systems aim to balance external information with the internal reasoning of the model, ensuring a more accurate and reliable output in clinical environments.
 
\label{sec:2understanding}
\section{Retrieval-augmented Generation in Vision}
\label{sec:Retrieval-augmented Generation in Vision}
\subsection{Image Generation}
\label{subsec:image_generation}

\begin{figure*}[ht!]
    \centering
    \includegraphics[width=0.99\textwidth]{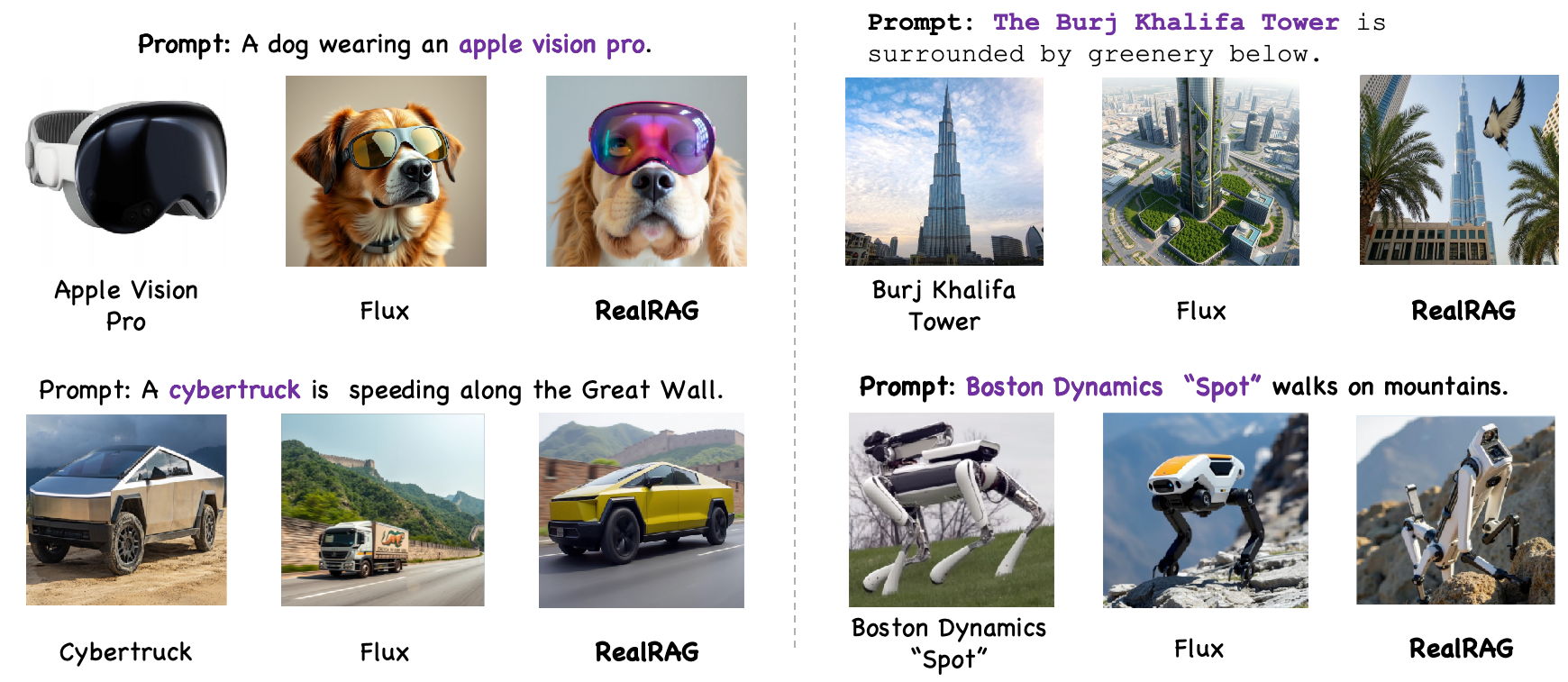}
    \caption{
    The visual results of unseen novel object generation from FLUX and RAG-based RealRAG~\cite{lyu2025realrag}.
    }    
    \label{fig:realrag_gen}
\end{figure*}

\begin{table*}[ht]
\centering
\setlength{\tabcolsep}{12pt}
\caption{Image Generation Methods with RAG techniques.}
\label{tab:imagegen}
\resizebox{\textwidth}{!}{  
\begin{tabular}{c|l|p{4cm}|p{4cm}|p{4cm}}
\midrule
Year & Method & Core Techniques & Retrieval Model & Challenges Addressed \\ \midrule
2022 & RA-Diffusion~\cite{blattmann2022retrieval} & Semi-parametric generative model & Nearest neighbor lookup & Efficient generation with fewer parameters \\ \midrule
2023 & Cioni~\cite{cioni2023diffusion} & Domain-specific database augmentation (\eg, artworks) & Zero-shot retrieval with CLIP pretrained on YFCC & Inaccurate generation in specialized domains, \eg, artwork \\ \midrule
2023 & ReMoDiffuse~\cite{zhang2023remodiffuse} & Diffusion-based motion generation with retrieval & Hybrid retrieval based on semantic / kinematic similarities & Improves motion generation diversity and generalization \\ \midrule
2024 & iRAG~\cite{iragbaidu2024} & - & Image retrieval integrated with generation & Solving "factuality hallucination" problem  \\ \midrule
2024 & FAI~\cite{wan2024factuality} & Fact-Augmented Intervention (FAI) using LLMs to incorporate factual data & GPT4 + ExpertQA~\cite{malaviya2023expertqa} & Nonfactual demographic generation, gender / racial composition issues \\ \midrule
2025 & ImageRAG~\cite{shalev2025imagerag} & Dynamic image retrieval with text prompts for contextual guidance & GPT4 + Cation then use text-image similarity metric for top-k samples & Reducing gap between generated and real-world images \\ \midrule
2025 & RealRAG~\cite{lyu2025realrag} & Self-reflective contrastive learning for fine-grained object generation & Reflective retriever & Enhances fine-grained object generation \\ \midrule
\end{tabular}
}
\end{table*}

\begin{figure}[t!]
    \centering
    \includegraphics[width=\linewidth]{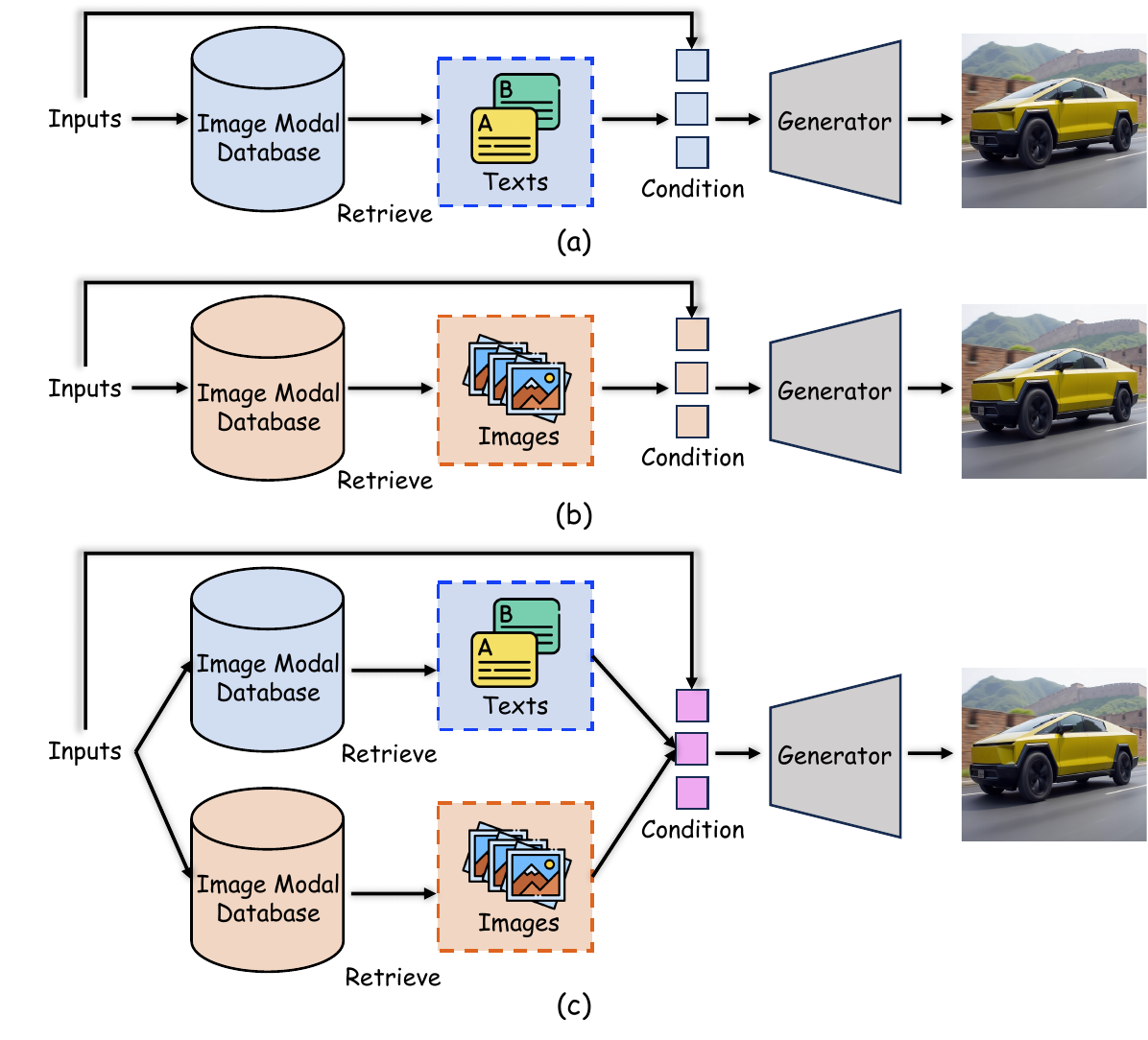}
    \caption{
    The overall of the retrieval-augmented image (video) generation frameworks. \textbf{(a)} Text-based framework; \textbf{(b)} Vision-based framework; \textbf{(c)} Dual-branch multimodal framework
    }
    \label{fig:image_generation_framework}
\end{figure}

Table~\ref{tab:imagegen} presents an overview of retrieval-augmented image generation methods, categorized based on their retrieval strategies and challenges addressed. As shown in Figure~\ref{fig:image_generation_framework}, these methods can be broadly classified into three categories based on the data modality used in their retrieval database:
\textbf{(1)} Text-based frameworks,
\textbf{(2)} Vision-based frameworks, and
\textbf{(3)} Multimodal frameworks.


\subsubsection{Text-based framework}
Prompt-based text-to-image generation relies on knowledge stored in the model's training memory to generate images. However, this approach often results in nonfactual demographic distributions, particularly when generating images that require specialized knowledge (\eg, historical images~\cite{wan2024factuality} and cultural images~\cite{cioni2023diffusion}). To address these issues, \cite{wan2024factuality} propose the Fact-Augmented Intervention (FAI), a method that instructs a Large Language Model (LLM) to reflect on verbalized or retrieved factual information regarding the gender and racial compositions of subjects in historical contexts, and to incorporate this information into the generation process of T2I models. Additionally, \cite{cioni2023diffusion} introduces a specialized domain-specific database (\eg, artworks) to augment the generation process.

\begin{figure*}[ht!]
    \centering
    \includegraphics[width=0.8\textwidth]{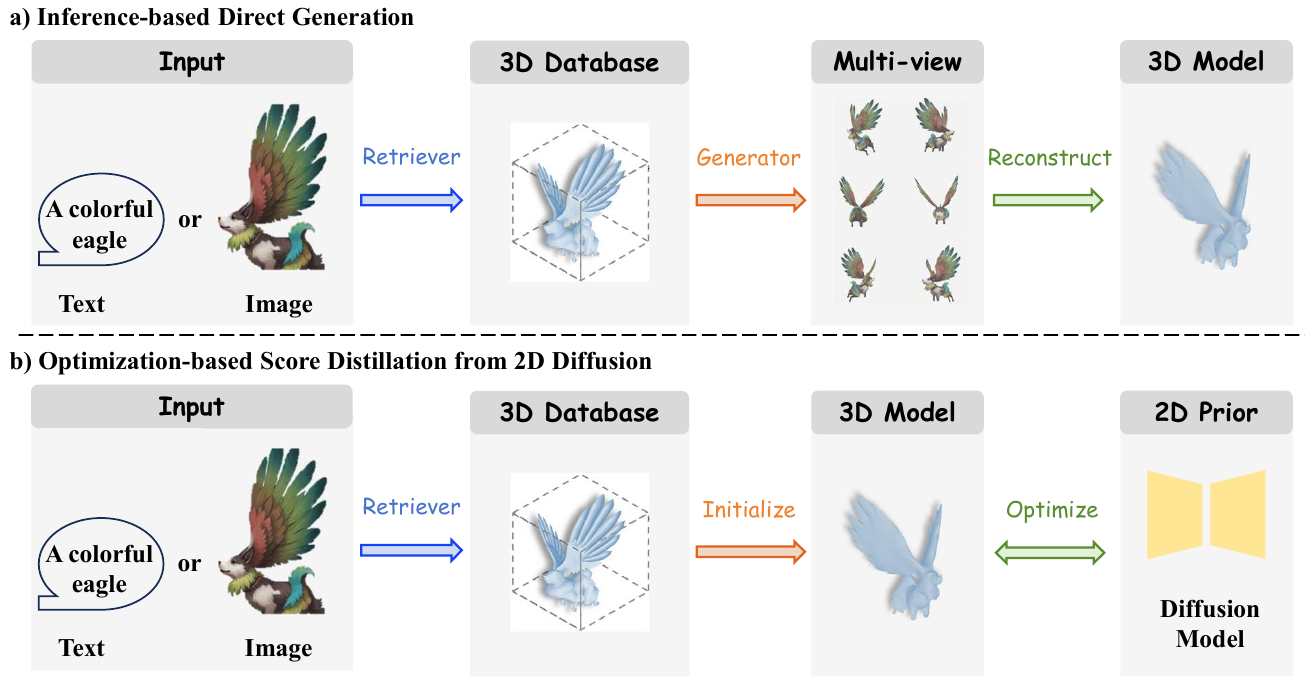}
    \caption{
    The overall pipeline of the retrieval-augmented 3D generation frameworks. Some images are borrowed from Phidias~\cite{wang2024phidias}.
    }
    \label{fig:3d_generation}
\end{figure*}
\subsubsection{Vision-based framework}
Recent advancements in novel architectures have significantly enhanced generative image synthesis, resulting in remarkable visual quality across various tasks. This success is largely attributed to the scalability of these architectures, which in turn has led to a dramatic increase in model complexity and the computational resources required for training. However, two main challenges persist: (1) As demonstrated in Figure~\ref{fig:image_generation_framework}, training pre-trained models in real-time remains difficult due to constraints posed by environmental factors, such as limited computational resources; and (2) the generated objects often appear too virtual, exhibiting a noticeable gap between them and real-world objects (Figure~\ref{fig:image_generation_framework}).

To address the first challenge, RA-Diffusion~\cite{blattmann2022retrieval} introduces a retrieval-augmented framework designed to create smaller generative models with access to a large image database. During training, the resulting semi-parametric generative models leverage this database through nearest neighbor lookups, eliminating the need to generate data "from scratch." Instead, they learn to compose new scenes based on retrieved visual instances. This approach not only improves generative performance with a reduced parameter count but also reduces computational requirements during training. RA-Diffusion further demonstrates the potential of generalizing to new knowledge in the form of alternative image databases, without necessitating additional training. 
Furthermore, ImageRAG~\cite{shalev2025imagerag} dynamically retrieves relevant images based on a given text prompt and incorporates them as contextual information to guide the generation process. In contrast, previous methods that leveraged retrieved images for enhancing generation typically required models to be specially trained for retrieval-based generation.
Following these approaches, various works have incorporated domain-specific image databases, such as garment~\cite{zhang2025garmentaligner}, traffic~\cite{ding2025realgen}, and layout~\cite{horita2024retrieval}. These methods pave the way for more efficient and sustainable generative models for image generation. However, these retrieval-augmented methods still overlook the challenge of improving the realism of generated images (mentioned in challenge 2).

As the scale of image generation models continues to expand, the scope of domains they can adapt to is also broadening. Consequently, the realism of the objects in generated images has become an increasingly important consideration. iRAG~\cite{iragbaidu2024} introduces the concept of "factuality hallucination" in visual generation models and addresses the issue by augmenting the realism of generated objects through image retrieval. iRAG highlights the limitations of large-scale image generation models and defines the "factuality hallucination" problem.
Furthermore, RealRAG~\cite{lyu2025realrag} achieves significant performance improvement in fine-grained and unseen object generation, which is powered by the proposed self-reflective contrastive learning. As shown in Figure~\ref{fig:realrag_gen}, the RealRAG demonstrates significant performance gain in fine-grained text-to-image generation.

\subsubsection{Multimodal Framework}
As illustrated in Figure~\ref{fig:image_generation_framework}, multimodal frameworks~\cite{zhang2023remodiffuse, yuan2025finerag} leverage diverse data sources from multimodal databases (\eg, MSCOCO~\cite{lin2014microsoft}, WebQA~\cite{chang2022webqa}) to improve visual generation.
ReMoDiffuse~\cite{zhang2023remodiffuse}, a diffusion-based motion generation model, incorporates retrieval to refine denoising. It introduces three components:
(1) Hybrid Retrieval, selecting database references via semantic and kinematic similarity;
(2) Semantic-Modulated Transformer, aligning retrieved information with target motion;
(3) Condition Mixture, optimizing database use during inference to mitigate classifier-free guidance sensitivity.
These frameworks enhance generative models by fusing multimodal features, improving completeness and diversity.

\subsection{Video Generation}
\label{subsec:video_generation}
The field of generative video synthesis has advanced significantly, with the integration of Retrieval-Augmented Generation (RAG) providing a promising solution to the challenges of creating high-quality, coherent, and contextually accurate videos. Recent works, such as Animate-AS~\cite{he2023animate}, explore the potential of leveraging pre-existing video assets in conjunction with generative models to enhance storytelling and video production efficiency. This approach utilizes two core modules: Motion Structure Retrieval and Structure-Guided Text-to-Video Synthesis. The first module retrieves video candidates based on text prompts, focusing on the motion structure of the retrieved clips, while the second module synthesizes new video content under the guidance of these retrieved structures. This technique allows for the efficient generation of videos with specific motions or layouts by reusing existing content and adjusting its appearance, rather than generating everything from scratch.

RAG-based video generation models represent a significant leap forward in generative video technology. By combining the advantages of retrieval systems and generative models, RAG offers a powerful framework for producing high-quality, personalized, and contextually accurate videos. With continued advancements in personalization, efficiency, and structural guidance, RAG methods hold immense potential for creative and commercial applications in video generation.

\begin{figure}[t!]
    \centering
    \includegraphics[width=1.0\linewidth]{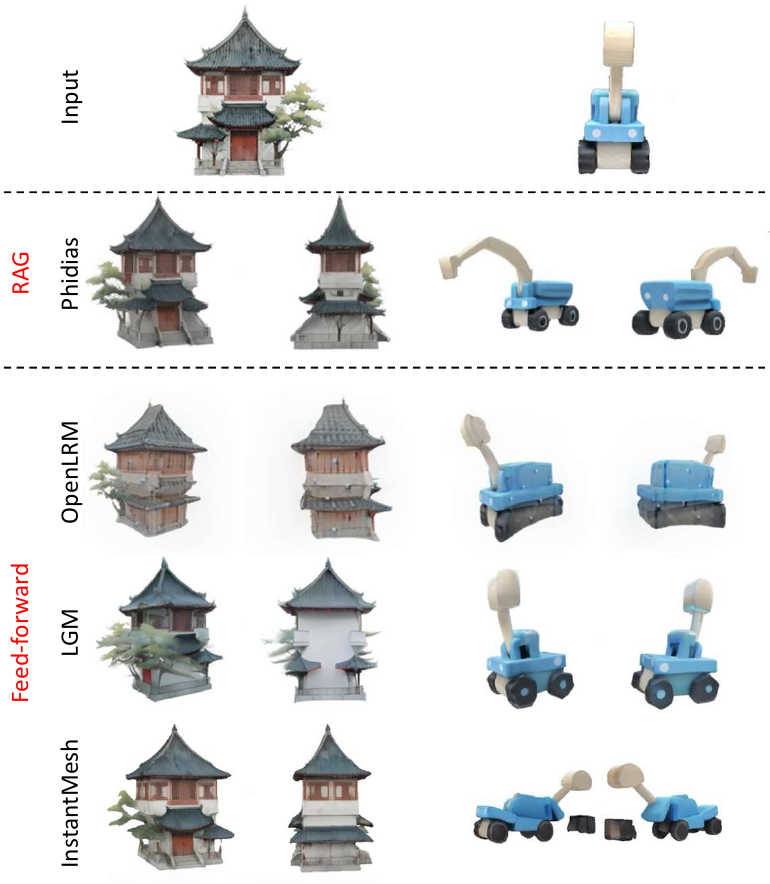}
    \caption{
    The comparison between RAG-based method Phidias~\cite{wang2024phidias} and Feed-forward methods, OpenLRM~\cite{hong2023lrm}, LGM~\cite{tang2024lgm}, and InstantMesh~\cite{xu2024instantmesh}. 
    }
    \label{fig:3d_comparison}
\end{figure}

\subsection{3D Generation}  
\label{subsec:3d_generation}  
Advancements in 2D diffusion models have significantly influenced progress in 3D generation. However, two key challenges persist in this field.  
The first challenge concerns data availability. Compared to large-scale 2D datasets, which contain billions of data pairs~\cite{schuhmann2021laion, schuhmann2022laion}, the largest 3D datasets~\cite{deitke2023objaverse, deitke2024objaverse} contain only around 10 million objects. Moreover, a substantial portion of these objects is excluded from training due to quality issues~\cite{xu2024instantmesh, tang2024edgerunner, chen2024meshxl}, limiting the performance of 3D generators relative to their 2D counterparts.  
The second challenge is generation quality. 3D generation is often constrained to text or single-image inputs, making it an ill-posed problem—i.e., the available conditions are insufficient to generate a fully accurate 3D model. This limitation restricts the realism and structural fidelity of generated 3D content, underscoring the need for improved conditioning mechanisms.

For 3D object generation, there are two mainstream paradigms that can be used to address these limitations. As shown in Figure~\ref{fig:3d_generation} a), Phidias~\cite{wang2024phidias} makes the early attempt and proposes a novel approach that utilizes a 3D database and leverages an RAG mechanism to enhance 3D generation. 
Phidias first retrieves a 3D reference model to render multi-view images, which are then used by a sparse-view-to-3D generator to reconstruct the 3D model.
As demonstrated in Figure~\ref{fig:3d_comparison}, the RAG-based method can generate more reasonable content compared with feed-forward methods~\cite{xu2024instantmesh, hong2023lrm, tang2024lgm}.
Moreover, as illustrated in Figure~\ref{fig:3d_generation} b), ReDream~\cite{seo2024retrieval} utilizes a retriever to find the most similar 3D object in the database, then optimizes this 3D object with a 2D diffusion prior.
Additionally, IRDiff~\cite{huanginteraction} has explored 3D molecular generation by integrating the diffusion model with RAG techniques.
ReMoDiffuse~\cite{Zhang_2023_ICCV} proposes to combine RAG into the field of 3D human motion generation.
Recently, Diorama~\cite{wu2024dioramaunleashingzeroshotsingleview} proposes to compose a scene by retrieving the components and placing them in the corresponding positions from some bounding boxes.
However, it is not followed by an optimization stage, making it hard to render a consistent image compared to the input image.

\label{sec:3generation}
\section{Retrieval-Augmented Generation in Embodied AI}
\label{sec:embodied_ai}

Retrieval-augmented generation (RAG) enhances Embodied AI by enabling agents to retrieve relevant external knowledge and generate informed actions or responses. This capability improves decision-making, adaptability, and overall performance in dynamic environments. As embodied agents grow more complex, integrating RAG strengthens task execution, multimodal perception, and domain-specific applications.

\subsection{RAG in Planning and Task Execution}
\label{subsec:embodied_planning}

Efficient task planning and execution are essential for intelligent agents operating in unpredictable environments. RAG enables agents to retrieve relevant knowledge in real time, improving adaptability and performance.
For instance, \cite{xu2024p} introduces a progressive retrieval model where agents continuously access external information to refine task execution. The system updates a database containing goal instructions, scene graphs, trajectory history, and task completion status after each interaction in environments like MINI-BEHAVIOR. This iterative learning process enhances task efficiency and adaptability.
Similarly, \cite{zhu2024retrieval} explores how RAG improves decision-making by retrieving external knowledge during task execution. This augmentation helps agents handle complex tasks and respond to unforeseen circumstances more effectively. These studies demonstrate how RAG enhances planning and execution, enabling agents to dynamically adjust to new contexts, a crucial capability for real-world applications.

\subsection{RAG in MultiModal Perception and Interaction}
\label{subsec:embodied_perception}

Embodied AI agents must process diverse sensory inputs—visual, auditory, and tactile—posing a challenge for effective perception and interaction. RAG facilitates the integration of multimodal data, improving agents' ability to understand and respond to their environment.
For example, \cite{nazar2024enwar} presents a framework that combines multimodal data, including wireless sensor inputs, with RAG to enhance situational awareness. By retrieving relevant contextual information, agents improve their ability to interpret complex environments and make informed decisions.
Additionally, \cite{xie2024embodied} proposes a method using non-parametric memory to store and retrieve sensory data, enhancing adaptability. By integrating RAG with memory systems, agents can better process past experiences and interact more effectively in dynamic settings. These advancements highlight RAG’s role in refining multimodal perception, enabling more natural and context-aware interactions.

\subsection{RAG in Specialized Domains}

\label{subsec:embodied_others}

RAG also offers significant benefits in specialized domains such as autonomous driving and traffic management, where agents must make real-time, context-aware decisions.
For instance, \cite{ding2024realgen} demonstrates how RAG retrieves real-time traffic data and historical patterns to optimize traffic flow, improving urban traffic management efficiency. Similarly, \cite{yuan2024rag} applies RAG to autonomous driving, enhancing explainability through retrieval-augmented in-context learning. By generating natural language explanations for driving decisions based on contextual demonstrations, the system increases transparency and trustworthiness. Notably, the model generalizes well to new environments, underscoring RAG’s effectiveness in real-world applications.
These examples illustrate how RAG enhances domain-specific decision-making, providing adaptability and transparency in areas like traffic optimization and autonomous systems.
Integrating RAG into Embodied AI significantly enhances agents' ability to execute tasks, perceive their environment, and interact naturally by leveraging external knowledge in real time. This improves adaptability and decision-making in dynamic environments, paving the way for advanced applications in domains such as autonomous driving and traffic management.

\begin{table*}[h!]
\centering
\setlength{\tabcolsep}{5pt}
\begin{tabular}{p{9cm}|p{9cm}}
\midrule
\textbf{Key Insights} & \textbf{Future Directions} \\ \midrule

\textbf{Mitigating Factuality Hallucination}  
Traditional generative models suffer from factual inconsistencies. \textit{iRAG}~\cite{iragbaidu2024} ensures factual alignment by retrieving real-world images. &  
\textbf{Real-Time and Dynamic Retrieval Systems}  
Future systems should integrate real-time retrieval from live databases (e.g., Wikipedia, news archives) to enhance factual alignment dynamically. \\ \midrule

\textbf{Enhancing Realism and Fine-Grained Details}  
\textit{RealRAG}~\cite{lyu2025realrag} improves object generation with fine details, producing high-fidelity images using self-reflective contrastive learning. &  
\textbf{Bias-Aware and Fairness-Enhanced Retrieval}  
Future methods should incorporate bias-detection pipelines to mitigate disparities in demographic representation (e.g., \textit{FAI}~\cite{wan2024factuality}). \\ \midrule

\textbf{Reducing Computational Overhead}  
Models like \textit{RA-Diffusion}~\cite{blattmann2022retrieval} reduce computational resources by retrieving images for efficient generation. &  
\textbf{Hybrid Multi-Scale Retrieval for Improved Realism}  
Combining low-level texture retrieval with high-level semantic retrieval could improve photorealism, especially for complex scenes. \\ \midrule

\textbf{Leveraging Multimodal Retrieval for Enriched Generation}  
Recent models like \textit{ReMoDiffuse}~\cite{zhang2023remodiffuse} integrate motion retrieval for dynamic generation. &  
\textbf{Federated and Distributed Retrieval for Privacy-Preserving Generation}  
Future work should explore decentralized federated retrieval models to protect user data privacy while pulling from diverse datasets. \\ \midrule

\textbf{Towards Adaptive and Context-Aware Generation}  
\textit{ImageRAG}~\cite{shalev2025imagerag} dynamically guides image generation based on contextual information, improving domain adaptability. &  
\textbf{Interactive and User-Adaptive Retrieval Mechanisms}  
Future systems should offer interactive retrieval interfaces, allowing users to modify parameters in real time for personalized outputs. \\ \midrule

\end{tabular}
\caption{Insights and Future Directions for RAG-based Image Generation}
\label{tab:insights_future}
\end{table*}

\section{Insights and New Outlook}
\label{sec:insights_outlook} 
\subsection{Insights and New Outlook for RAU}
\label{subsec:insights_outlook_RAU} 
\subsubsection{RAU in Video Understanding}
Retrieval-augmented techniques have advanced multimodal video retrieval and understanding by integrating external knowledge to enhance contextual comprehension. However, current methods face persistent challenges in balancing computational efficiency with temporal reasoning. While retrieval mechanisms have improved contextual grounding, they still struggle with efficiently processing long videos and maintaining coherence across extended sequences.
Recent innovations such as arbitrary-length video processing and token reduction techniques have improved efficiency, but challenges remain in areas like memory constraints and long-range dependency modeling. Additionally, maintaining narrative consistency, visual coherence, and security in retrieval-based video applications remains an open challenge.

The future directions could be: (1) Dynamic Resource Allocation: Future systems should optimize retrieval efficiency by prioritizing critical video segments without compromising comprehension.  
and (2) Enhanced Memory Management: Improving memory efficiency in retrieval-augmented video models is essential for processing long sequences while maintaining semantic coherence.  

\subsubsection{RAU in Multimodal Learning}
Incorporating external knowledge into vision-language models has proven effective for multimodal tasks. However, large-scale vision-language models, such as the \textit{LLaVA} series, remain prone to hallucinations when encountering novel knowledge. Multimodal RAG (\textit{mRAG}) methods offer a partial solution by retrieving relevant external information to improve factual alignment.
Despite progress in \textit{mRAG}-based models, challenges persist in balancing modality interactions in multi-task settings and avoiding over-reliance on text retrieval. Some models disproportionately prioritize textual information while underutilizing visual inputs, leading to imbalances in cross-modal understanding. Additionally, the overlap between training datasets and retrieved external data can impact both fairness and retrieval effectiveness.

The future directions could be:

\noindent \textbf{Optimized Multimodal Sampling:}  
Enhancing content selection and integration across modalities can improve robustness in multi-task settings.  

\noindent \textbf{Advanced Multimodal Fusion:}  
Better balancing modality contributions can prevent over-reliance on textual cues.  

\noindent \textbf{Cross-Task Generalization:}  
Developing retrieval-based models that transfer knowledge across tasks will enhance broader applicability.

RAU continues to reshape pattern recognition, video analysis, and multimodal learning by integrating external knowledge sources to enhance model performance. While recent advancements have addressed challenges related to factual grounding, efficiency, and cross-modal interactions, future research must focus on developing \textbf{\textit{adaptive, scalable, and secure retrieval strategies}} to further improve generalization and reliability in real-world applications.

\subsection{Insights and New Outlook for RAG}
\label{subsec:insights_outlook_RAG} 
\subsubsection{RAG for Image Generation}
By leveraging external databases to retrieve relevant information, RAG-based methods address several longstanding challenges in generative models, including factuality hallucination, domain adaptation, computational efficiency, and realism in generated images. 

We summarize key insights from recent advancements and promising future directions in retrieval-augmented image generation in the table labeled Table~\ref{tab:insights_future}. Despite recent progress, several challenges remain, and the table presents these insights and future directions in a structured manner.

\subsubsection{RAG in 3D Content Generation}
The integration of RAG mechanisms into 3D object- and scene-level generation presents significant potential for future research. One of the primary challenges in this domain is the scarcity of high-quality 3D object datasets, a limitation that extends to 3D scene generation. Current approaches in scene synthesis often rely on component composition~\cite{bai2023componerf} and scene graph representations~\cite{dhamo2021graph}. Incorporating RAG into 3D scene generation could enhance diversity and quality by leveraging external knowledge sources to guide the synthesis process.

Advancements in 2D diffusion models have driven progress in 3D generation, yet two major challenges persist:

\noindent \textbf{Data Availability:} Compared to large-scale 2D datasets containing billions of image-text pairs~\cite{schuhmann2021laion, schuhmann2022laion}, the largest 3D datasets~\cite{deitke2023objaverse, deitke2024objaverse} contain only around 10 million objects. Moreover, a significant portion of these objects is excluded from training due to quality concerns~\cite{xu2024instantmesh, tang2024edgerunner, chen2024meshxl}, limiting the performance of 3D generative models relative to their 2D counterparts.

\noindent \textbf{Generation Quality:} 3D generation remains an inherently ill-posed problem, as it is typically conditioned on text or single-image inputs, which provide insufficient information to fully reconstruct a detailed 3D model. This constraint affects the realism and structural fidelity of generated content, highlighting the need for enhanced conditioning mechanisms that incorporate richer external knowledge sources.

By addressing these challenges, the integration of RAG into 3D content generation could significantly improve data efficiency and synthesis quality, opening new avenues for research and practical applications.

\subsection{RAG in Embodied AI}  
\label{subsec:insights_outlook_Embodied_AI} 
As shown in Table~\ref{tab:insights_future_embodiedAI}, we list the insights and future directions of RAG in embodied AI.
RAG has significantly advanced Embodied AI by integrating external knowledge retrieval into decision-making, enhancing task execution, multimodal perception, and specialized applications.
Despite these advancements, challenges remain in optimizing retrieval efficiency, improving cross-modal reasoning, and scaling real-world applications.

\begin{table*}[ht]
\centering
\renewcommand{\arraystretch}{1.3}
\setlength{\tabcolsep}{5pt}
\begin{tabular}{p{9cm}|p{9cm}}
\midrule
\textbf{Key Insights} & \textbf{Future Directions} \\ \midrule

\textbf{Adaptive Decision-Making in Dynamic Environments}  
RAG-enabled systems, such as those in \cite{xu2024p}, retrieve relevant task knowledge in real time, allowing agents to adapt to novel challenges and optimize performance. &  
\textbf{Optimized Real-Time Retrieval}  
Efficient retrieval pipelines are essential for real-time applications in robotics and autonomous systems. Future research should focus on \textbf{low-latency retrieval mechanisms} that balance efficiency with accuracy. \\ \midrule

\textbf{Enhanced Multimodal Perception and Interaction}  
By leveraging RAG, frameworks such as \cite{nazar2024enwar} and \cite{xie2024embodied} improve perception accuracy and enable more natural interactions. &  
\textbf{Scalable Multimodal Fusion}  
Improved \textbf{multimodal fusion techniques} are needed for seamless integration of retrieved information, including \textbf{hierarchical retrieval architectures} for dynamic cross-modal reasoning. \\ \midrule

\textbf{Optimized Decision-Making in Specialized Domains}  
Systems like \cite{ding2024realgen} and \cite{yuan2024rag} improve decision accuracy and explainability, making AI-driven systems more reliable. &  
\textbf{Context-Aware and Personalized Retrieval}  
Future RAG-enhanced Embodied AI should implement \textbf{adaptive retrieval mechanisms} tailored to task-specific and environmental contexts. \\ \midrule

\textbf{Memory-Augmented Planning and Execution}  
Progressive retrieval techniques, as shown in \cite{xu2024p}, improve adaptability and efficiency by dynamically refining action plans. &  
\textbf{Privacy-Preserving Retrieval}  
Future RAG frameworks should incorporate \textbf{privacy-aware retrieval} to ensure data security and regulatory compliance, especially in sensitive applications like personal assistants and medical robotics. \\ \midrule
\end{tabular}
\caption{Insights and Future Directions for RAG in Embodied AI}
\label{tab:insights_future_embodiedAI}
\end{table*}

\subsection{Application}  
\label{subsec:application} 

\subsubsection{Multimodal Interaction and Understanding}
Multimodal interaction and understanding represent a significant application domain for computer vision, involving the comprehension of multiple modalities such as text, images, and videos. By integrating information from different modalities, RAG models can achieve a better understanding of complex scenes and tasks. For instance, in Chatbot applications using visual question answering (VQA)~\cite{chen2022murag}, RAG enhances the ability to generate accurate and contextually relevant responses. Additionally, RAG has been effectively applied to various downstream tasks~\cite{liu2023learning}, such as image classification, object detection, and segmentation, to enhance model performance and adaptability through external knowledge retrieval.

\subsubsection{Vision Generation Tasks}
Visual generation tasks aim to enhance the models' generative capabilities by retrieving external knowledge, thereby producing high-quality visual content. These tasks encompass not only generating images or videos from textual descriptions but also creating 3D models. The applications of generation tasks are extensive, with existing uses including providing inspiration for creative design and artistic creation~\cite{cioni2023diffusion}. Potential future applications could involve creating immersive experiences in virtual and augmented reality, as well as generating personalized multimedia content in content creation to meet the diverse needs of users.

\subsubsection{Multifaceted Applications of Embodied AI}
Embodied AI, which integrates perception, interaction, and action capabilities, enables agents to interact with the physical world. RAG plays a crucial role in embodied AI by retrieving external knowledge in real time, significantly enhancing the agents' interaction and decision-making capabilities. RAG allows agents to dynamically adjust their strategies, and optimize task planning and execution. These enhancements make embodied AI agents more adaptable and efficient, paving the way for practical applications in areas such as autonomous driving~\cite{ding2024realgen} and intelligent robotics~\cite{xu2024p}.

\subsubsection{Domain-Specific Applications}
The domain-specific applications of RAG span a variety of fields, with notable contributions in medical and industrial contexts. Within the medical realm, RAG retrieves medical databases to ensure the accuracy of generated medical reports and diagnoses~\cite{xia2024rule}.  Meanwhile, in the industrial field, it digs into relevant defect patterns and standards to sharpen the precision and efficiency of quality control processes~\cite{2025stx}. These applications highlight the versatility and power of RAG across diverse domains, illustrating its promising potential usage in specific domains.
\section{Conclusion}

In this survey, we have presented a comprehensive review of retrieval-augmented generation (RAG) techniques in the context of computer vision (CV). We explored the integration of RAG into visual understanding, visual generation, and embodied vision, highlighting the significant potential for enhancing tasks such as object recognition, scene parsing, and 3D content generation. By leveraging external knowledge, RAG improves model performance, addressing challenges related to knowledge limitations, outdated information, and domain-specific expertise.
We have identified key advancements in RAG for CV, including its application in multimodal understanding and its role in improving the efficiency and quality of visual generation tasks. Despite its potential, RAG in CV still faces challenges, such as retrieval efficiency, modality alignment, and computational cost, which need to be addressed for broader adoption. Our work also outlines future research directions, particularly in real-time retrieval optimization, cross-modal fusion, and the integration of RAG into embodied AI, 3D content generation, and robotics.
This survey serves as a foundation for future exploration of retrieval-augmented techniques in computer vision, offering valuable insights into how these methods can be applied to enhance real-world decision-making and model adaptability. We hope our work encourages further research that will continue to push the boundaries of RAG's applications in CV, leading to more robust, intelligent, and adaptable vision systems.


\bibliographystyle{IEEEtran}  
\bibliography{main}

\begin{thebibliography}{100}
\providecommand{\url}[1]{#1}
\csname url@samestyle\endcsname
\providecommand{\newblock}{\relax}
\providecommand{\bibinfo}[2]{#2}
\providecommand{\BIBentrySTDinterwordspacing}{\spaceskip=0pt\relax}
\providecommand{\BIBentryALTinterwordstretchfactor}{4}
\providecommand{\BIBentryALTinterwordspacing}{\spaceskip=\fontdimen2\font plus
\BIBentryALTinterwordstretchfactor\fontdimen3\font minus \fontdimen4\font\relax}
\providecommand{\BIBforeignlanguage}[2]{{%
\expandafter\ifx\csname l@#1\endcsname\relax
\typeout{** WARNING: IEEEtran.bst: No hyphenation pattern has been}%
\typeout{** loaded for the language `#1'. Using the pattern for}%
\typeout{** the default language instead.}%
\else
\language=\csname l@#1\endcsname
\fi
#2}}
\providecommand{\BIBdecl}{\relax}
\BIBdecl

\bibitem{gao2023retrieval}
Y.~Gao, Y.~Xiong, X.~Gao, K.~Jia, J.~Pan, Y.~Bi, Y.~Dai, J.~Sun, and H.~Wang, ``Retrieval-augmented generation for large language models: A survey,'' \emph{arXiv preprint arXiv:2312.10997}, 2023.

\bibitem{fan2024surveyragmeetingllms}
W.~Fan, Y.~Ding, L.~Ning, S.~Wang, H.~Li, D.~Yin, T.~Chua, and Q.~Li, ``A survey on {RAG} meeting llms: Towards retrieval-augmented large language models,'' in \emph{Proceedings of the 30th {ACM} {SIGKDD} Conference on Knowledge Discovery and Data Mining, {KDD} 2024, Barcelona, Spain, August 25-29, 2024}, R.~Baeza{-}Yates and F.~Bonchi, Eds.\hskip 1em plus 0.5em minus 0.4em\relax {ACM}, 2024, pp. 6491--6501.

\bibitem{hu2024rag}
Y.~Hu and Y.~Lu, ``Rag and rau: A survey on retrieval-augmented language model in natural language processing,'' \emph{arXiv preprint arXiv:2404.19543}, 2024.

\bibitem{zhao2024retrieval}
L.~Zhao, X.~Chen, E.~Z. Chen, Y.~Liu, T.~Chen, and S.~Sun, ``Retrieval-augmented few-shot medical image segmentation with foundation models,'' \emph{arXiv preprint arXiv:2408.08813}, 2024.

\bibitem{gupta2024comprehensive}
S.~Gupta, R.~Ranjan, and S.~N. Singh, ``A comprehensive survey of retrieval-augmented generation (rag): Evolution, current landscape and future directions,'' \emph{arXiv preprint arXiv:2410.12837}, 2024.

\bibitem{zhao2024retrievalaigc}
P.~Zhao, H.~Zhang, Q.~Yu, Z.~Wang, Y.~Geng, F.~Fu, L.~Yang, W.~Zhang, J.~Jiang, and B.~Cui, ``Retrieval-augmented generation for ai-generated content: A survey,'' \emph{arXiv preprint arXiv:2402.19473}, 2024.

\bibitem{yu2024evaluation}
H.~Yu, A.~Gan, K.~Zhang, S.~Tong, Q.~Liu, and Z.~Liu, ``Evaluation of retrieval-augmented generation: A survey,'' in \emph{CCF Conference on Big Data}.\hskip 1em plus 0.5em minus 0.4em\relax Springer, 2024, pp. 102--120.

\bibitem{procko2024graph}
T.~T. Procko and O.~Ochoa, ``Graph retrieval-augmented generation for large language models: A survey,'' in \emph{2024 Conference on AI, Science, Engineering, and Technology (AIxSET)}.\hskip 1em plus 0.5em minus 0.4em\relax IEEE, 2024, pp. 166--169.

\bibitem{zhou2024trustworthiness}
Y.~Zhou, Y.~Liu, X.~Li, J.~Jin, H.~Qian, Z.~Liu, C.~Li, Z.~Dou, T.-Y. Ho, and P.~S. Yu, ``Trustworthiness in retrieval-augmented generation systems: A survey,'' \emph{arXiv preprint arXiv:2409.10102}, 2024.

\bibitem{singh2025agentic}
A.~Singh, A.~Ehtesham, S.~Kumar, and T.~T. Khoei, ``Agentic retrieval-augmented generation: A survey on agentic rag,'' \emph{arXiv preprint arXiv:2501.09136}, 2025.

\bibitem{ni2025towards}
B.~Ni, Z.~Liu, L.~Wang, Y.~Lei, Y.~Zhao, X.~Cheng, Q.~Zeng, L.~Dong, Y.~Xia, K.~Kenthapadi \emph{et~al.}, ``Towards trustworthy retrieval augmented generation for large language models: A survey,'' \emph{arXiv preprint arXiv:2502.06872}, 2025.

\bibitem{lewis2020retrieval}
P.~Lewis, E.~Perez, A.~Piktus, F.~Petroni, V.~Karpukhin, N.~Goyal, H.~K{\"u}ttler, M.~Lewis, W.-t. Yih, T.~Rockt{\"a}schel \emph{et~al.}, ``Retrieval-augmented generation for knowledge-intensive nlp tasks,'' \emph{NeurIPS}, vol.~33, pp. 9459--9474, 2020.

\bibitem{yu2024visrag}
S.~Yu, C.~Tang, B.~Xu, J.~Cui, J.~Ran, Y.~Yan, Z.~Liu, S.~Wang, X.~Han, Z.~Liu \emph{et~al.}, ``Visrag: Vision-based retrieval-augmented generation on multi-modality documents,'' \emph{arXiv preprint arXiv:2410.10594}, 2024.

\bibitem{he2016deep}
K.~He, X.~Zhang, S.~Ren, and J.~Sun, ``Deep residual learning for image recognition,'' in \emph{CVPR}, 2016, pp. 770--778.

\bibitem{lyu2024unibind}
Y.~Lyu, X.~Zheng, J.~Zhou, and L.~Wang, ``Unibind: Llm-augmented unified and balanced representation space to bind them all,'' in \emph{CVPR}, 2024, pp. 26\,752--26\,762.

\bibitem{zhang2022dino}
H.~Zhang, F.~Li, S.~Liu, L.~Zhang, H.~Su, J.~Zhu, L.~M. Ni, and H.-Y. Shum, ``Dino: Detr with improved denoising anchor boxes for end-to-end object detection,'' \emph{arXiv preprint arXiv:2203.03605}, 2022.

\bibitem{ravi2024sam}
N.~Ravi, V.~Gabeur, Y.-T. Hu, R.~Hu, C.~Ryali, T.~Ma, H.~Khedr, R.~R{\"a}dle, C.~Rolland, L.~Gustafson \emph{et~al.}, ``Sam 2: Segment anything in images and videos,'' \emph{arXiv preprint arXiv:2408.00714}, 2024.

\bibitem{liu2024rar}
Z.~Liu, Z.~Sun, Y.~Zang, W.~Li, P.~Zhang, X.~Dong, Y.~Xiong, D.~Lin, and J.~Wang, ``Rar: Retrieving and ranking augmented mllms for visual recognition,'' \emph{arXiv preprint arXiv:2403.13805}, 2024.

\bibitem{wang2024phidias}
Z.~Wang, T.~Wang, Z.~He, G.~Hancke, Z.~Liu, and R.~W. Lau, ``Phidias: A generative model for creating 3d content from text, image, and 3d conditions with reference-augmented diffusion,'' \emph{arXiv preprint arXiv:2409.11406}, 2024.

\bibitem{jeong2025videorag}
S.~Jeong, K.~Kim, J.~Baek, and S.~J. Hwang, ``Videorag: Retrieval-augmented generation over video corpus,'' 2025.

\bibitem{miech2019howto100m}
A.~Miech, D.~Zhukov, J.-B. Alayrac, M.~Tapaswi, I.~Laptev, and J.~Sivic, ``Howto100m: Learning a text-video embedding by watching hundred million narrated video clips,'' in \emph{ICCV}, 2019, pp. 2630--2640.

\bibitem{Arefeen2024ViTA}
M.~A. Arefeen, B.~Debnath, M.~Y.~S. Uddin, and S.~Chakradhar, ``Vita: An efficient video-to-text algorithm using vlm for rag-based video analysis system,'' in \emph{CVPRW}, 2024.

\bibitem{piadyk2023streetaware}
Y.~Piadyk, J.~Rulff, E.~Brewer, M.~Hosseini, K.~Ozbay, M.~Sankaradas, S.~Chakradhar, and C.~Silva, ``Streetaware: A high-resolution synchronized multimodal urban scene dataset,'' \emph{Sensors}, vol.~23, no.~7, p. 3710, 2023.

\bibitem{kossmann2023extract}
F.~Kossmann, Z.~Wu, E.~Lai, N.~Tatbul, L.~Cao, T.~Kraska, and S.~Madden, ``Extract-transform-load for video streams,'' \emph{arXiv preprint arXiv:2310.04830}, 2023.

\bibitem{zhang2024omagent}
L.~Zhang, T.~Zhao, H.~Ying, Y.~Ma, and K.~Lee, ``Omagent: A multi-modal agent framework for complex video understanding with task divide-and-conquer,'' in \emph{arxiv}, 2024.

\bibitem{wang2024sok}
A.~Wang, B.~Wu, S.~Chen, Z.~Chen, H.~Guan, W.-N. Lee, L.~E. Li, and C.~Gan, ``Sok-bench: A situated video reasoning benchmark with aligned open-world knowledge,'' in \emph{CVPR}, 2024, pp. 13\,384--13\,394.

\bibitem{sankaradas2025streamingragrealtimecontextualretrieval}
M.~Sankaradas, R.~K. Rajendran, and S.~T. Chakradhar, ``Streamingrag: Real-time contextual retrieval and generation framework,'' arXiv, 2025.

\bibitem{long2022retrieval}
A.~Long, W.~Yin, T.~Ajanthan, V.~Nguyen, P.~Purkait, R.~Garg, A.~Blair, C.~Shen, and A.~van~den Hengel, ``Retrieval augmented classification for long-tail visual recognition,'' in \emph{CVPR}, 2022, pp. 6959--6969.

\bibitem{kim2024retrieval}
J.~Kim, E.~Cho, S.~Kim, and H.~J. Kim, ``Retrieval-augmented open-vocabulary object detection,'' in \emph{CVPR}, 2024, pp. 17\,427--17\,436.

\bibitem{zhou2023style}
Y.~Zhou and G.~Long, ``Style-aware contrastive learning for multi-style image captioning,'' in \emph{EACL}, 2023.

\bibitem{li2024evcap}
J.~Li, D.~M. Vo, A.~Sugimoto, and H.~Nakayama, ``Evcap: Retrieval-augmented image captioning with external visual-name memory for open-world comprehension,'' in \emph{CVPR}, 2024, pp. 13\,733--13\,742.

\bibitem{ramos2023retrieval}
R.~Ramos, D.~Elliott, and B.~Martins, ``Retrieval-augmented image captioning,'' \emph{arXiv preprint arXiv:2302.08268}, 2023.

\bibitem{li2024understanding}
W.~Li, J.~Li, R.~Ramos, R.~Tang, and D.~Elliott, ``Understanding retrieval robustness for retrieval-augmented image captioning,'' \emph{arXiv preprint arXiv:2406.02265}, 2024.

\bibitem{yang2023mm}
Z.~Yang, L.~Li, J.~Wang, K.~Lin, E.~Azarnasab, F.~Ahmed, Z.~Liu, C.~Liu, M.~Zeng, and L.~Wang, ``Mm-react: Prompting chatgpt for multimodal reasoning and action,'' in \emph{arxiv}, 2023.

\bibitem{jin2024flashrag}
J.~Jin, Y.~Zhu, X.~Yang, C.~Zhang, and Z.~Dou, ``Flashrag: A modular toolkit for efficient retrieval-augmented generation research,'' in \emph{arXiv}, 2024.

\bibitem{bonomo2025visual}
M.~Bonomo and S.~Bianco, ``Visual rag: Expanding mllm visual knowledge without fine-tuning,'' 2025.

\bibitem{xue2024enhanced}
J.~Xue, Q.~Deng, F.~Yu, Y.~Wang, J.~Wang, and Y.~Li, ``Enhanced multimodal rag-llm for accurate visual question answering,'' 2024.

\bibitem{kim2024vipcap}
T.~Kim, S.~Lee, S.-W. Kim, and D.-J. Kim, ``Vipcap: Retrieval text-based visual prompts for lightweight image captioning,'' 2024.

\bibitem{huang2024vinci}
Y.~Huang, J.~Xu, B.~Pei, Y.~He, G.~Chen, L.~Yang, X.~Chen, Y.~Wang, Z.~Nie, J.~Liu \emph{et~al.}, ``Vinci: A real-time embodied smart assistant based on egocentric vision-language model,'' arXiv, 2024.

\bibitem{xiong2025streaming}
H.~Xiong, Z.~Yang, J.~Yu, Y.~Zhuge, L.~Zhang, J.~Zhu, and H.~Lu, ``Streaming video understanding and multi-round interaction with memory-enhanced knowledge,'' 2025.

\bibitem{luo2024video}
Y.~Luo, X.~Zheng, X.~Yang, G.~Li, H.~Lin, J.~Huang, J.~Ji, F.~Chao, J.~Luo, and R.~Ji, ``Video-rag: Visually-aligned retrieval-augmented long video comprehension,'' in \emph{arXiv}, 2024.

\bibitem{arefeen2024irag}
M.~A. Arefeen, B.~Debnath, M.~Y.~S. Uddin, and S.~Chakradhar, ``irag: Advancing rag for videos with an incremental approach,'' in \emph{CIKM}, 2024.

\bibitem{spolaor2020systematic}
N.~Spola\~or, H.~D. Lee, W.~S.~R. Takaki, L.~A. Ensina, C.~S.~R. Coy, and F.~C. Wu, ``A systematic review on content-based video retrieval,'' in \emph{Eng. Appl. Artif. Intell.}, 2020.

\bibitem{chen2024large}
Y.~Chen, S.~Guo, and L.~Wang, ``A large-scale study on video action dataset condensation,'' 2024.

\bibitem{yu2024cross}
X.~Yu, X.~Feng, Y.~Li, M.~Liao, Y.-Q. Yu, X.~Feng, W.~Zhong, R.~Chen, M.~Hu, and J.~Wu, ``Cross-lingual text-rich visual comprehension: An information theory perspective,'' in \emph{arxiv}, 2024.

\bibitem{he2023animate}
Y.~He, M.~Xia, H.~Chen, X.~Cun, Y.~Gong, J.~Xing, Y.~Zhang, X.~Wang, C.~Weng, Y.~Shan \emph{et~al.}, ``Animate-a-story: Storytelling with retrieval-augmented video generation,'' \emph{arXiv}, 2023.

\bibitem{zhang2024dialogue}
M.~Zhang, Z.~Wang, L.~Chen, K.~Liu, and J.~Lin, ``Dialogue director: Bridging the gap in dialogue visualization for multimodal storytelling,'' \emph{arXiv}, 2024.

\bibitem{wen2024ensemble}
W.~Wen, Y.~Wang, N.~Birkbeck, and B.~Adsumilli, ``An ensemble approach to short-form video quality assessment using multimodal llm,'' 2024.

\bibitem{fang2025retrievals}
H.~Fang, X.~Sui, H.~Yu, J.~Kong, S.~Yu, B.~Chen, H.~Wu, and S.-T. Xia, ``Retrievals can be detrimental: A contrastive backdoor attack paradigm on retrieval-augmented diffusion models,'' 2025.

\bibitem{hong2024free}
F.-T. Hong, Z.~Xu, H.~Liu, Q.~Lin, L.~Song, Z.~Shu, Y.~Zhou, D.~Ceylan, and D.~Xu, ``Free-viewpoint human animation with pose-correlated reference selection,'' \emph{arXiv}, 2024.

\bibitem{luo2025graphbasedcrossdomainknowledgedistillation}
B.~Luo, J.~Wang, Z.~Wang, J.~Zhu, and X.~Zhao, ``Graph-based cross-domain knowledge distillation for cross-dataset text-to-image person retrieval,'' arXiv, 2025.

\bibitem{xu2025zero}
Y.~Xu, Y.~Sun, B.~Zhai, M.~Li, W.~Liang, Y.~Li, and S.~Du, ``Zero-shot video moment retrieval via off-the-shelf multimodal large language models,'' 2025.

\bibitem{xu2024multi}
Y.~Xu, Y.~Sun, B.~Zhai, Z.~Xie, Y.~Jia, and S.~Du, ``Multi-modal fusion and query refinement network for video moment retrieval and highlight detection,'' in \emph{ICME}, 2024.

\bibitem{chen2024human}
C.~Chen, F.~Lv, Y.~Guan, P.~Wang, S.~Yu, Y.~Zhang, and Z.~Tang, ``Human-guided image generation for expanding small-scale training image datasets,'' 2024.

\bibitem{goyal2017makingvvqa}
Y.~Goyal, T.~Khot, D.~Summers{-}Stay, D.~Batra, and D.~Parikh, ``Making the {V} in {VQA} matter: Elevating the role of image understanding in visual question answering,'' in \emph{2017 {IEEE} Conference on Computer Vision and Pattern Recognition, {CVPR} 2017, Honolulu, HI, USA, July 21-26, 2017}.\hskip 1em plus 0.5em minus 0.4em\relax {IEEE} Computer Society, 2017, pp. 6325--6334.

\bibitem{gurari2018vizwiz}
D.~Gurari, Q.~Li, A.~J. Stangl, A.~Guo, C.~Lin, K.~Grauman, J.~Luo, and J.~P. Bigham, ``Vizwiz grand challenge: Answering visual questions from blind people,'' in \emph{2018 {IEEE} Conference on Computer Vision and Pattern Recognition, {CVPR} 2018, Salt Lake City, UT, USA, June 18-22, 2018}.\hskip 1em plus 0.5em minus 0.4em\relax {IEEE} Computer Society, 2018, pp. 3608--3617.

\bibitem{marino2019okvqa}
K.~Marino, M.~Rastegari, A.~Farhadi, and R.~Mottaghi, ``{OK-VQA:} {A} visual question answering benchmark requiring external knowledge,'' in \emph{{IEEE} Conference on Computer Vision and Pattern Recognition, {CVPR} 2019, Long Beach, CA, USA, June 16-20, 2019}.\hskip 1em plus 0.5em minus 0.4em\relax Computer Vision Foundation / {IEEE}, 2019, pp. 3195--3204.

\bibitem{10.1007/978-3-031-20074-8_9}
D.~Schwenk, A.~Khandelwal, C.~Clark, K.~Marino, and R.~Mottaghi, ``A-okvqa: A benchmark for visual question answering using world knowledge,'' in \emph{Computer Vision -- ECCV 2022}, S.~Avidan, G.~Brostow, M.~Ciss{\'e}, G.~M. Farinella, and T.~Hassner, Eds.\hskip 1em plus 0.5em minus 0.4em\relax Cham: Springer Nature Switzerland, 2022, pp. 146--162.

\bibitem{lu2022learnexplainmultimodal}
P.~Lu, S.~Mishra, T.~Xia, L.~Qiu, K.~Chang, S.~Zhu, O.~Tafjord, P.~Clark, and A.~Kalyan, ``Learn to explain: Multimodal reasoning via thought chains for science question answering,'' in \emph{NeurIPS 35: Annual Conference on Neural Information Processing Systems 2022, NeurIPS 2022, New Orleans, LA, USA, November 28 - December 9, 2022}, S.~Koyejo, S.~Mohamed, A.~Agarwal, D.~Belgrave, K.~Cho, and A.~Oh, Eds., 2022.

\bibitem{hu2023opendomain}
H.~Hu, Y.~Luan, Y.~Chen, U.~Khandelwal, M.~Joshi, K.~Lee, K.~Toutanova, and M.~Chang, ``Open-domain visual entity recognition: Towards recognizing millions of wikipedia entities,'' in \emph{{IEEE/CVF} International Conference on Computer Vision, {ICCV} 2023, Paris, France, October 1-6, 2023}.\hskip 1em plus 0.5em minus 0.4em\relax {IEEE}, 2023, pp. 12\,031--12\,041.

\bibitem{chen2023pretrainedvision}
Y.~Chen, H.~Hu, Y.~Luan, H.~Sun, S.~Changpinyo, A.~Ritter, and M.-W. Chang, ``Can pre-trained vision and language models answer visual information-seeking questions?'' in \emph{Proceedings of the 2023 Conference on Empirical Methods in Natural Language Processing}, H.~Bouamor, J.~Pino, and K.~Bali, Eds.\hskip 1em plus 0.5em minus 0.4em\relax Singapore: ACL, 2023, pp. 14\,948--14\,968.

\bibitem{cho2024m3docrag}
J.~Cho, D.~Mahata, O.~Irsoy, Y.~He, and M.~Bansal, ``M3docrag: Multi-modal retrieval is what you need for multi-page multi-document understanding,'' 2024.

\bibitem{faysse2024colpali}
M.~Faysse, H.~Sibille, T.~Wu, B.~Omrani, G.~Viaud, C.~HUDELOT, and P.~Colombo, ``Colpali: Efficient document retrieval with vision language models,'' in \emph{ICLR}, 2025.

\bibitem{dong2025mmdocir}
K.~Dong, Y.~Chang, X.~D. Goh, D.~Li, R.~Tang, and Y.~Liu, ``Mmdocir: Benchmarking multi-modal retrieval for long documents,'' 2025.

\bibitem{10.1007/978-3-319-10602-1_48}
T.-Y. Lin, M.~Maire, S.~Belongie, J.~Hays, P.~Perona, D.~Ramanan, P.~Doll{\'a}r, and C.~L. Zitnick, ``Microsoft coco: Common objects in context,'' in \emph{Computer Vision -- ECCV 2014}, D.~Fleet, T.~Pajdla, B.~Schiele, and T.~Tuytelaars, Eds.\hskip 1em plus 0.5em minus 0.4em\relax Cham: Springer International Publishing, 2014, pp. 740--755.

\bibitem{gupta2019lvisdataset}
A.~Gupta, P.~Doll{\'{a}}r, and R.~B. Girshick, ``{LVIS:} {A} dataset for large vocabulary instance segmentation,'' in \emph{{IEEE} Conference on Computer Vision and Pattern Recognition, {CVPR} 2019, Long Beach, CA, USA, June 16-20, 2019}.\hskip 1em plus 0.5em minus 0.4em\relax Computer Vision Foundation / {IEEE}, 2019, pp. 5356--5364.

\bibitem{wang2023v3det}
J.~Wang, P.~Zhang, T.~Chu, Y.~Cao, Y.~Zhou, T.~Wu, B.~Wang, C.~He, and D.~Lin, ``V3det: Vast vocabulary visual detection dataset,'' in \emph{{IEEE/CVF} International Conference on Computer Vision, {ICCV} 2023, Paris, France, October 1-6, 2023}.\hskip 1em plus 0.5em minus 0.4em\relax {IEEE}, 2023, pp. 19\,787--19\,797.

\bibitem{li2022elevater}
C.~Li, H.~Liu, L.~H. Li, P.~Zhang, J.~Aneja, J.~Yang, P.~Jin, H.~Hu, Z.~Liu, Y.~J. Lee, and J.~Gao, ``{ELEVATER:} {A} benchmark and toolkit for evaluating language-augmented visual models,'' in \emph{NeurIPS 35: Annual Conference on Neural Information Processing Systems 2022, NeurIPS 2022, New Orleans, LA, USA, November 28 - December 9, 2022}, S.~Koyejo, S.~Mohamed, A.~Agarwal, D.~Belgrave, K.~Cho, and A.~Oh, Eds., 2022.

\bibitem{liu2024mmbench}
Y.~Liu, H.~Duan, Y.~Zhang, B.~Li, S.~Zhang, W.~Zhao, Y.~Yuan, J.~Wang, C.~He, Z.~Liu, K.~Chen, and D.~Lin, ``Mmbench: Is your multi-modal model an all-around player?'' 2023.

\bibitem{LiGGWWZS24}
B.~Li, Y.~Ge, Y.~Ge, G.~Wang, R.~Wang, R.~Zhang, and Y.~Shan, ``Seed-bench: Benchmarking multimodal large language models,'' in \emph{CVPR}, 2024, pp. 13\,299--13\,308.

\bibitem{chen2024rightway}
L.~Chen, J.~Li, X.~Dong, P.~Zhang, Y.~Zang, Z.~Chen, H.~Duan, J.~Wang, Y.~Qiao, D.~Lin, and F.~Zhao, ``Are we on the right way for evaluating large vision-language models?'' in \emph{NeurIPS, NeurIPS 2024, Vancouver, BC, Canada, December 10 - 15, 2024}, 2024.

\bibitem{wu2025visualrag}
Y.~Wu, Q.~Long, J.~Li, J.~Yu, and W.~Wang, ``Visual-rag: Benchmarking text-to-image retrieval augmented generation for visual knowledge intensive queries,'' 2025.

\bibitem{hu2025mragbench}
W.~Hu, J.-C. Gu, Z.-Y. Dou, M.~Fayyaz, P.~Lu, K.-W. Chang, and N.~Peng, ``{MRAG}-bench: Vision-centric evaluation for retrieval-augmented multimodal models,'' in \emph{ICLR}, 2025.

\bibitem{wasserman2025realmmra}
N.~Wasserman, R.~Pony, O.~Naparstek, A.~R. Goldfarb, E.~Schwartz, U.~Barzelay, and L.~Karlinsky, ``Real-mm-rag: A real-world multi-modal retrieval benchmark,'' 2025.

\bibitem{li2023evaluatingobject}
Y.~Li, Y.~Du, K.~Zhou, J.~Wang, X.~Zhao, and J.-R. Wen, ``Evaluating object hallucination in large vision-language models,'' in \emph{Proceedings of the 2023 Conference on Empirical Methods in Natural Language Processing}, H.~Bouamor, J.~Pino, and K.~Bali, Eds.\hskip 1em plus 0.5em minus 0.4em\relax Singapore: ACL, 2023, pp. 292--305.

\bibitem{mortaheb2025rag}
M.~Mortaheb, M.~A.~A. Khojastepour, S.~T. Chakradhar, and S.~Ulukus, ``Rag-check: Evaluating multimodal retrieval augmented generation performance,'' \emph{arXiv}, 2025.

\bibitem{chen2022murag}
W.~Chen, H.~Hu, X.~Chen, P.~Verga, and W.~Cohen, ``{M}u{RAG}: Multimodal retrieval-augmented generator for open question answering over images and text,'' in \emph{EMNLP}, Y.~Goldberg, Z.~Kozareva, and Y.~Zhang, Eds.\hskip 1em plus 0.5em minus 0.4em\relax Abu Dhabi, United Arab Emirates: ACL, 2022, pp. 5558--5570.

\bibitem{liu2023learning}
H.~Liu, K.~Son, J.~Yang, C.~Liu, J.~Gao, Y.~J. Lee, and C.~Li, ``Learning customized visual models with retrieval-augmented knowledge,'' in \emph{CVPR}.\hskip 1em plus 0.5em minus 0.4em\relax {IEEE}, 2023, pp. 15\,148--15\,158.

\bibitem{lin2014microsoft}
T.-Y. Lin, M.~Maire, S.~Belongie, J.~Hays, P.~Perona, D.~Ramanan, P.~Doll{\'a}r, and C.~L. Zitnick, ``Microsoft coco: Common objects in context,'' in \emph{Computer vision--ECCV 2014: 13th European conference, zurich, Switzerland, September 6-12, 2014, proceedings, part v 13}.\hskip 1em plus 0.5em minus 0.4em\relax Springer, 2014, pp. 740--755.

\bibitem{Goyal2019VQA}
Y.~Goyal, T.~Khot, D.~Summers{-}Stay, D.~Batra, and D.~Parikh, ``Making the {V} in {VQA} matter: Elevating the role of image understanding in visual question answering,'' in \emph{CVPR}.\hskip 1em plus 0.5em minus 0.4em\relax {IEEE} Computer Society, 2017, pp. 6325--6334.

\bibitem{li2024search}
C.~Li, Z.~Li, C.~Jing, S.~Liu, W.~Shao, Y.~Wu, P.~Luo, Y.~Qiao, and K.~Zhang, ``Search{LVLM}s: A plug-and-play framework for augmenting large vision-language models by searching up-to-date internet knowledge,'' in \emph{NeurIPS}, 2024.

\bibitem{lin2023finegrained}
W.~Lin, J.~Chen, J.~Mei, A.~Coca, and B.~Byrne, ``Fine-grained late-interaction multi-modal retrieval for retrieval augmented visual question answering,'' in \emph{NeurIPS}, 2023.

\bibitem{tan2024retrieval}
C.~Tan, J.~Wei, L.~Sun, Z.~Gao, S.~Li, B.~Yu, R.~Guo, and S.~Z. Li, ``Retrieval meets reasoning: Even high-school textbook knowledge benefits multimodal reasoning,'' \emph{CoRR}, vol. abs/2405.20834, 2024.

\bibitem{chen2024mllm}
Z.~Chen, C.~Xu, Y.~Qi, and J.~Guo, ``Mllm is a strong reranker: Advancing multimodal retrieval-augmented generation via knowledge-enhanced reranking and noise-injected training,'' 2024.

\bibitem{mathew2021docvqa}
M.~Mathew, D.~Karatzas, and C.~V. Jawahar, ``Docvqa: A dataset for vqa on document images,'' in \emph{WACV}, 2021, pp. 2199--2208.

\bibitem{ragflow}
RAGFlow, ``Ragflow is an open-source rag (retrieval-augmented generation) engine based on deep document understanding,'' 2024.

\bibitem{riedler2024text}
M.~Riedler and S.~Langer, ``Beyond text: Optimizing rag with multimodal inputs for industrial applications,'' 2024.

\bibitem{ye2023mplugdoc}
J.~Ye, A.~Hu, H.~Xu, Q.~Ye, M.~Yan, Y.~Dan, C.~Zhao, G.~Xu, C.~Li, J.~Tian, Q.~Qi, J.~Zhang, and F.~Huang, ``mplug-docowl: Modularized multimodal large language model for document understanding,'' 2023.

\bibitem{wang2024charxiv}
Z.~Wang, M.~Xia, L.~He, H.~Chen, Y.~Liu, R.~Zhu, K.~Liang, X.~Wu, H.~Liu, S.~Malladi, A.~Chevalier, S.~Arora, and D.~Chen, ``Charxiv: Charting gaps in realistic chart understanding in multimodal {LLM}s,'' in \emph{NeurIPS}, 2024.

\bibitem{hu2023reveal}
Z.~Hu, A.~Iscen, C.~Sun, Z.~Wang, K.~Chang, Y.~Sun, C.~Schmid, D.~A. Ross, and A.~Fathi, ``Reveal: Retrieval-augmented visual-language pre-training with multi-source multimodal knowledge memory,'' in \emph{CVPR}.\hskip 1em plus 0.5em minus 0.4em\relax {IEEE}, 2023, pp. 23\,369--23\,379.

\bibitem{rao2023retrieval}
J.~Rao, Z.~Shan, L.~Liu, Y.~Zhou, and Y.~Yang, ``Retrieval-based knowledge augmented vision language pre-training,'' in \emph{ACM MM}.\hskip 1em plus 0.5em minus 0.4em\relax {ACM}, 2023, pp. 5399--5409.

\bibitem{rao2024raven}
V.~N. Rao, S.~Choudhary, A.~Deshpande, R.~K. Satzoda, and S.~Appalaraju, ``Raven: Multitask retrieval augmented vision-language learning,'' 2024.

\bibitem{sun2024surf}
J.~Sun, J.~Zhang, Y.~Zhou, Z.~Su, X.~Qu, and Y.~Cheng, ``{SUR}f: Teaching large vision-language models to selectively utilize retrieved information,'' in \emph{EMNLP}, Y.~Al-Onaizan, M.~Bansal, and Y.-N. Chen, Eds.\hskip 1em plus 0.5em minus 0.4em\relax Miami, Florida, USA: ACL, 2024, pp. 7611--7629.

\bibitem{qi2024roravlm}
J.~Qi, Z.~Xu, R.~Shao, Y.~Chen, J.~Di, Y.~Cheng, Q.~Wang, and L.~Huang, ``Rora-vlm: Robust retrieval-augmented vision language models,'' 2024.

\bibitem{xia2024mmed}
P.~Xia, K.~Zhu, H.~Li, T.~Wang, W.~Shi, S.~Wang, L.~Zhang, J.~Zou, and H.~Yao, ``Mmed-rag: Versatile multimodal rag system for medical vision language models,'' \emph{arXiv preprint arXiv:2410.13085}, 2024.

\bibitem{xia2024rule}
P.~Xia, K.~Zhu, H.~Li, H.~Zhu, Y.~Li, G.~Li, L.~Zhang, and H.~Yao, ``Rule: Reliable multimodal rag for factuality in medical vision language models,'' in \emph{EMNLP}, 2024, pp. 1081--1093.

\bibitem{lyu2025realrag}
Y.~Lyu, X.~Zheng, L.~Jiang, Y.~Yan, X.~Zou, H.~Zhou, L.~Zhang, and X.~Hu, ``Realrag: Retrieval-augmented realistic image generation via self-reflective contrastive learning,'' \emph{arXiv preprint arXiv:2502.00848}, 2025.

\bibitem{blattmann2022retrieval}
A.~Blattmann, R.~Rombach, K.~Oktay, J.~M{\"u}ller, and B.~Ommer, ``Retrieval-augmented diffusion models,'' \emph{NeurIPS}, vol.~35, pp. 15\,309--15\,324, 2022.

\bibitem{cioni2023diffusion}
D.~Cioni, L.~Berlincioni, F.~Becattini, and A.~Del~Bimbo, ``Diffusion based augmentation for captioning and retrieval in cultural heritage,'' in \emph{ICCV}, 2023, pp. 1707--1716.

\bibitem{zhang2023remodiffuse}
M.~Zhang, X.~Guo, L.~Pan, Z.~Cai, F.~Hong, H.~Li, L.~Yang, and Z.~Liu, ``Remodiffuse: Retrieval-augmented motion diffusion model,'' in \emph{ICCV}, 2023, pp. 364--373.

\bibitem{iragbaidu2024}
``Image-based rag: \url{https://baike.baidu.com/item/iRAG/65102065?fr=ge_ala},'' 2024.

\bibitem{wan2024factuality}
Y.~Wan, D.~Wu, H.~Wang, and K.-W. Chang, ``The factuality tax of diversity-intervened text-to-image generation: Benchmark and fact-augmented intervention,'' \emph{arXiv preprint arXiv:2407.00377}, 2024.

\bibitem{malaviya2023expertqa}
C.~Malaviya, S.~Lee, S.~Chen, E.~Sieber, M.~Yatskar, and D.~Roth, ``Expertqa: Expert-curated questions and attributed answers,'' \emph{arXiv preprint arXiv:2309.07852}, 2023.

\bibitem{shalev2025imagerag}
R.~Shalev-Arkushin, R.~Gal, A.~H. Bermano, and O.~Fried, ``Imagerag: Dynamic image retrieval for reference-guided image generation,'' \emph{arXiv preprint arXiv:2502.09411}, 2025.

\bibitem{zhang2025garmentaligner}
S.~Zhang, Z.~Chong, X.~Zhang, H.~Li, Y.~Cheng, Y.~Yan, and X.~Liang, ``Garmentaligner: Text-to-garment generation via retrieval-augmented multi-level corrections,'' in \emph{ECCV}.\hskip 1em plus 0.5em minus 0.4em\relax Springer, 2025, pp. 148--164.

\bibitem{ding2025realgen}
W.~Ding, Y.~Cao, D.~Zhao, C.~Xiao, and M.~Pavone, ``Realgen: Retrieval augmented generation for controllable traffic scenarios,'' in \emph{ECCV}.\hskip 1em plus 0.5em minus 0.4em\relax Springer, 2025, pp. 93--110.

\bibitem{horita2024retrieval}
D.~Horita, N.~Inoue, K.~Kikuchi, K.~Yamaguchi, and K.~Aizawa, ``Retrieval-augmented layout transformer for content-aware layout generation,'' in \emph{CVPR}, 2024, pp. 67--76.

\bibitem{yuan2025finerag}
H.~Yuan, Z.~Zhao, S.~Wang, S.~Xiao, M.~Ni, Z.~Liu, and Z.~Dou, ``Finerag: Fine-grained retrieval-augmented text-to-image generation,'' in \emph{Proceedings of the 31st International Conference on Computational Linguistics}, 2025, pp. 11\,196--11\,205.

\bibitem{chang2022webqa}
Y.~Chang, M.~Narang, H.~Suzuki, G.~Cao, J.~Gao, and Y.~Bisk, ``Webqa: Multihop and multimodal qa,'' in \emph{CVPR}, 2022, pp. 16\,495--16\,504.

\bibitem{hong2023lrm}
Y.~Hong, K.~Zhang, J.~Gu, S.~Bi, Y.~Zhou, D.~Liu, F.~Liu, K.~Sunkavalli, T.~Bui, and H.~Tan, ``Lrm: Large reconstruction model for single image to 3d,'' \emph{arXiv preprint arXiv:2311.04400}, 2023.

\bibitem{tang2024lgm}
J.~Tang, Z.~Chen, X.~Chen, T.~Wang, G.~Zeng, and Z.~Liu, ``Lgm: Large multi-view gaussian model for high-resolution 3d content creation,'' in \emph{ECCV}.\hskip 1em plus 0.5em minus 0.4em\relax Springer, 2024, pp. 1--18.

\bibitem{xu2024instantmesh}
J.~Xu, W.~Cheng, Y.~Gao, X.~Wang, S.~Gao, and Y.~Shan, ``Instantmesh: Efficient 3d mesh generation from a single image with sparse-view large reconstruction models,'' \emph{arXiv preprint arXiv:2404.07191}, 2024.

\bibitem{schuhmann2021laion}
C.~Schuhmann, R.~Vencu, R.~Beaumont, R.~Kaczmarczyk, C.~Mullis, A.~Katta, T.~Coombes, J.~Jitsev, and A.~Komatsuzaki, ``Laion-400m: Open dataset of clip-filtered 400 million image-text pairs,'' \emph{arXiv preprint arXiv:2111.02114}, 2021.

\bibitem{schuhmann2022laion}
C.~Schuhmann, R.~Beaumont, R.~Vencu, C.~Gordon, R.~Wightman, M.~Cherti, T.~Coombes, A.~Katta, C.~Mullis, M.~Wortsman \emph{et~al.}, ``Laion-5b: An open large-scale dataset for training next generation image-text models,'' \emph{NeurIPS}, vol.~35, pp. 25\,278--25\,294, 2022.

\bibitem{deitke2023objaverse}
M.~Deitke, D.~Schwenk, J.~Salvador, L.~Weihs, O.~Michel, E.~VanderBilt, L.~Schmidt, K.~Ehsani, A.~Kembhavi, and A.~Farhadi, ``Objaverse: A universe of annotated 3d objects,'' in \emph{CVPR}, 2023, pp. 13\,142--13\,153.

\bibitem{deitke2024objaverse}
M.~Deitke, R.~Liu, M.~Wallingford, H.~Ngo, O.~Michel, A.~Kusupati, A.~Fan, C.~Laforte, V.~Voleti, S.~Y. Gadre \emph{et~al.}, ``Objaverse-xl: A universe of 10m+ 3d objects,'' \emph{NeurIPS}, vol.~36, 2024.

\bibitem{tang2024edgerunner}
J.~Tang, Z.~Li, Z.~Hao, X.~Liu, G.~Zeng, M.-Y. Liu, and Q.~Zhang, ``Edgerunner: Auto-regressive auto-encoder for artistic mesh generation,'' \emph{arXiv preprint arXiv:2409.18114}, 2024.

\bibitem{chen2024meshxl}
S.~Chen, X.~Chen, A.~Pang, X.~Zeng, W.~Cheng, Y.~Fu, F.~Yin, Y.~Wang, Z.~Wang, C.~Zhang \emph{et~al.}, ``Meshxl: Neural coordinate field for generative 3d foundation models,'' \emph{arXiv preprint arXiv:2405.20853}, 2024.

\bibitem{seo2024retrieval}
J.~Seo, S.~Hong, W.~Jang, I.~H. Kim, M.~Kwak, D.~Lee, and S.~Kim, ``Retrieval-augmented score distillation for text-to-3d generation,'' \emph{arXiv preprint arXiv:2402.02972}, 2024.

\bibitem{huanginteraction}
Z.~Huang, L.~Yang, X.~Zhou, C.~Qin, Y.~Yu, X.~Zheng, Z.~Zhou, W.~Zhang, Y.~Wang, and W.~Yang, ``Interaction-based retrieval-augmented diffusion models for protein-specific 3d molecule generation,'' in \emph{ICML}.

\bibitem{Zhang_2023_ICCV}
M.~Zhang, X.~Guo, L.~Pan, Z.~Cai, F.~Hong, H.~Li, L.~Yang, and Z.~Liu, ``Remodiffuse: Retrieval-augmented motion diffusion model,'' in \emph{ICCV (ICCV)}, October 2023, pp. 364--373.

\bibitem{wu2024dioramaunleashingzeroshotsingleview}
Q.~Wu, D.~Iliash, D.~Ritchie, M.~Savva, and A.~X. Chang, ``Diorama: Unleashing zero-shot single-view 3d scene modeling,'' 2024.

\bibitem{xu2024p}
W.~Xu, M.~Wang, W.~Zhou, and H.~Li, ``P-rag: Progressive retrieval augmented generation for planning on embodied everyday task,'' in \emph{Proceedings of the 32nd ACM International Conference on Multimedia}, 2024, pp. 6969--6978.

\bibitem{zhu2024retrieval}
Y.~Zhu, Z.~Ou, X.~Mou, and J.~Tang, ``Retrieval-augmented embodied agents,'' in \emph{CVPR}, 2024, pp. 17\,985--17\,995.

\bibitem{nazar2024enwar}
A.~M. Nazar, A.~Celik, M.~Y. Selim, A.~Abdallah, D.~Qiao, and A.~M. Eltawil, ``Enwar: A rag-empowered multi-modal llm framework for wireless environment perception,'' \emph{arXiv preprint arXiv:2410.18104}, 2024.

\bibitem{xie2024embodied}
Q.~Xie, S.~Y. Min, T.~Zhang, K.~Xu, A.~Bajaj, R.~Salakhutdinov, M.~Johnson-Roberson, and Y.~Bisk, ``Embodied-rag: General non-parametric embodied memory for retrieval and generation,'' in \emph{Language Gamification-NeurIPS 2024 Workshop}, 2024.

\bibitem{ding2024realgen}
W.~Ding, Y.~Cao, D.~Zhao, C.~Xiao, and M.~Pavone, ``Realgen: Retrieval augmented generation for controllable traffic scenarios,'' in \emph{ECCV}.\hskip 1em plus 0.5em minus 0.4em\relax Springer, 2024, pp. 93--110.

\bibitem{yuan2024rag}
J.~Yuan, S.~Sun, D.~Omeiza, B.~Zhao, P.~Newman, L.~Kunze, and M.~Gadd, ``Rag-driver: Generalisable driving explanations with retrieval-augmented in-context learning in multi-modal large language model,'' \emph{arXiv preprint arXiv:2402.10828}, 2024.

\bibitem{bai2023componerf}
H.~Bai, Y.~Lyu, L.~Jiang, S.~Li, H.~Lu, X.~Lin, and L.~Wang, ``Componerf: Text-guided multi-object compositional nerf with editable 3d scene layout,'' \emph{arXiv preprint arXiv:2303.13843}, 2023.

\bibitem{dhamo2021graph}
H.~Dhamo, F.~Manhardt, N.~Navab, and F.~Tombari, ``Graph-to-3d: End-to-end generation and manipulation of 3d scenes using scene graphs,'' in \emph{ICCV}, 2021, pp. 16\,352--16\,361.

\bibitem{2025stx}
``Retrieval augmented generation: 40\% reduction in inspection times and notable improvement in defect detection rates,'' 2024.

\end{thebibliography}

\end{document}